\documentclass[lettersize,journal]{IEEEtran}
\usepackage{amsmath,amsfonts}
\usepackage{algorithmic}
\usepackage{algorithm}
\usepackage{array}
\usepackage[caption=false,font=normalsize,labelfont=sf,textfont=sf]{subfig}
\usepackage{textcomp}
\usepackage{stfloats}
\usepackage{url}
\usepackage{verbatim}
\usepackage{graphicx}
\usepackage{cite}
\usepackage{soul, color, xcolor}

\usepackage[colorlinks,
linkcolor=blue,
anchorcolor=blue,
citecolor=blue,backref]{hyperref}

\usepackage{caption}

\usepackage{multirow}
\usepackage{colortbl}
\usepackage[normalem]{ulem}
\useunder{\uline}{\ul}{}

\usepackage{amssymb}
\begin{document}

\title{SiamSeg: Self-Training with Contrastive Learning for Unsupervised Domain Adaptation Semantic Segmentation in Remote Sensing}

\author{
	Bin Wang, Fei Deng, Shuang Wang, 
	Wen Luo, ~\IEEEmembership{Member,~IEEE}, 
	Zhixuan Zhang, 
	Peifan Jiang, ~\IEEEmembership{Graduate Student Member,~IEEE}

	\thanks{\emph{(Corresponding author: Fei Deng)}
}

	\thanks{Bin Wang, Fei Deng and Wen Luo are with the College of Computer Science and Cyber Security, Chengdu University of Technology, Chengdu 610059, China (e-mail: woldier@foxmail.com; luowen0724@qq.com; dengfei@cdut.edu.cn).
	
	Shuang Wang and Peifan Jiang are with the Key Laboratory of Earth Exploration and Information Techniques of Ministry of Education and College of Geophysics, Chengdu University of Technology, Chengdu 610059, China (e-mail: wangs@stu.cdut.edu.cn; jpeifan@qq.com).

	Zhixuan Zhang is with College of Mechanical and Vehicle Engineering, Changchun University, Changchun, 629100, China (e-mail: 2046236458@qq.com;). 
	
	}

}

\markboth{Journal of \LaTeX\ Class Files,~Vol.~14, No.~8, August~2021}%
{Shell \MakeLowercase{\textit{et al.}}: A Sample Article Using IEEEtran.cls for IEEE Journals}


\maketitle

\begin{abstract}

Semantic segmentation of remote sensing (RS) images is a challenging yet essential task with broad applications. While deep learning, particularly supervised learning with large-scale labeled datasets, has significantly advanced this field, the acquisition of high-quality labeled data remains costly and time-intensive. Unsupervised domain adaptation (UDA) provides a promising alternative by enabling models to learn from unlabeled target domain data while leveraging labeled source domain data. Recent self-training (ST) approaches employing pseudo-label generation have shown potential in mitigating domain discrepancies. However, the application of ST to RS image segmentation remains underexplored. Factors such as variations in ground sampling distance, imaging equipment, and geographic diversity exacerbate domain shifts, limiting model performance across domains. In that case, existing ST methods, due to significant domain shifts in cross-domain RS images, often underperform.
To address these challenges, we propose integrating contrastive learning into UDA, enhancing the model’s ability to capture semantic information in the target domain by maximizing the similarity between augmented views of the same image. This additional supervision improves the model’s representational capacity and segmentation performance in the target domain. Extensive experiments conducted on RS datasets, including Potsdam, Vaihingen, and LoveDA, demonstrate that our method, SimSeg, outperforms existing approaches, achieving state-of-the-art results. Visualization and quantitative analyses further validate SimSeg’s superior ability to learn from the target domain. The code is publicly available at
\url{https://github.com/woldier/SiamSeg}.
\end{abstract}


\section{Introduction}
\begin{figure}[!t]
	\centering
	\includegraphics[width=0.75\linewidth]{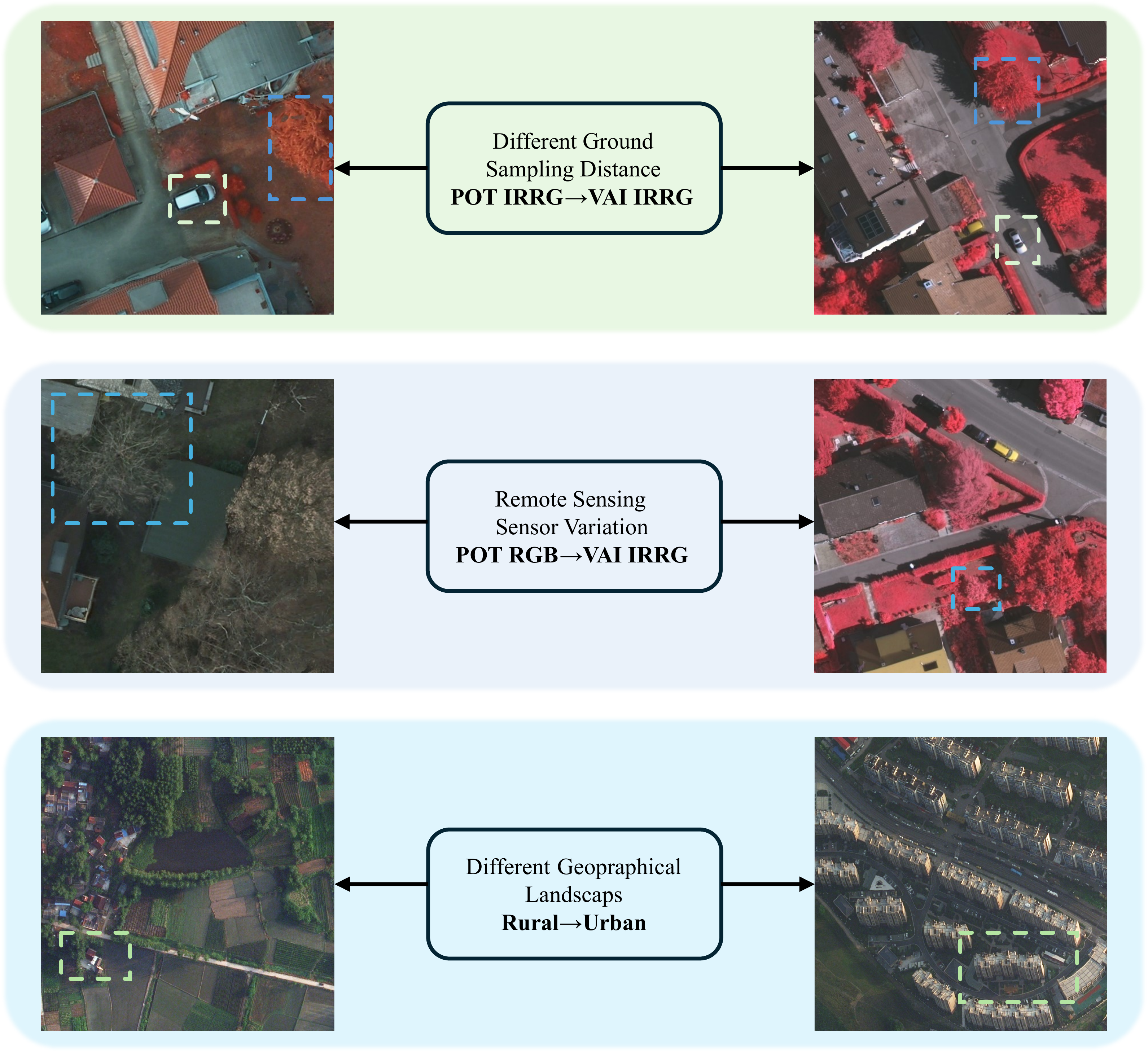}
	\caption{The main challenges in the task of cross-domain semantic segmentation of remote sensing images. 
		These challenges include the problem of domain bias due to ground sampling distances, sensor differences, and variations in geographic landscapes, which affect the model's ability to generalize across different datasets. 
		Understanding these domain shift issues is crucial for improving the accuracy and robustness of semantic segmentation of RS images.}
	\label{fig1}
\end{figure}
\IEEEPARstart{R}{emote} sensing techniques are widely employed in various visual tasks, including RS images classification \cite{xia2017,cheng2017,Geng2024,Zhang2024}, object detection \cite{xia2018,lin2020,Qiu2024}, and semantic segmentation \cite{lyu2020,hou2022,deng2023,mao2023,luo2024,Nie2023}. Among these, the semantic segmentation task aims to accurately classify each pixel in remote sensing images for pixel-level target recognition. The extensive application of remote sensing image segmentation in urban planning, flood control, and environmental monitoring \cite{li2022} has garnered increasing attention from researchers, prompting deeper exploration of the topic.

In recent years, deep learning-based semantic segmentation methods have made significant strides, leading to the emergence of many top-performing models, such as Fully Convolutional Neural Networks \cite{long2015,ronneberger2015,zhao2017} and Transformer-Based Models \cite{zheng2021,xie2021,xu2021}. However, the effectiveness of these methods heavily depends on the distributional similarity between the training and test data. When a domain shift occurs between different datasets, model performance significantly deteriorates. In practice, this domain shift problem is particularly pronounced due to the diversity of geographic regions, imaging conditions, and equipment used in remote sensing datasets, resulting in insufficient generalization capability of existing methods. 

To address the domain shift problem and establish effective associations between source and target domains, cross-domain semantic segmentation of remote sensing images has emerged as a significant research direction. Unsupervised domain adaptation (UDA), a subset of transfer learning, aims to tackle the generalization challenge when the source domain has labeled data while the target domain contains only unlabeled data. Existing UDA methods can be broadly categorized into two groups: adversarial learning based methods and self-training (ST) based methods \cite{toldo2020}.
Adversarial learning based approaches assist segmentation networks in minimizing the differences between the distributions of feature maps in the source and target domains by introducing a discriminator  \cite{liu2022,wilson2020,hu2022,Xu2023}. 
Unlike adversarial learning and generative domain adaptation methods, self-training (ST) methods \cite{li2022} do not rely on additional discriminators. The ST strategy facilitates cross-domain knowledge transfer by generating pseudo-label  \cite{lee2013,zou2019,zou2018,tranheden2021,chen2021}. 

Although many classical UDA methods have been successfully applied to natural scenes, the domain shift problem in remote sensing images is more complex, as illustrated in Fig. \ref{fig1}, stemming from factors such as ground sampling distances, sensor type discrepancies, and geographic landscape variations. This results in a larger domain gap for cross-domain RS images and significantly degrades the performance of methods that work well in natural scenes when directly applied to remote sensing data. 

Directly applying the ST method to cross-domain RS image semantic segmentation does not capture the feature information of the target domain image well, which leads to the performance degradation of the ST method in the target domain.
The rise of contrast learning in computer vision demonstrates its powerful capability to capture semantic information in images without relying on labeled data, resulting in enhanced feature representation. This addresses the issue of a large domain gap, which prevents the application of ST methods to learn the target domain effectively through pseudo-label. Based on this observation, this paper proposes SiamSeg, which introduces contrast learning to the unsupervised domain adaptation task of semantic segmentation in remote sensing images. Leveraging the robust feature representation capability of contrast learning, SiamSeg effectively addresses the insufficient semantic information learning caused by the weak supervisory signals of pseudo-label in the target domain, significantly enhancing segmentation network performance.

\begin{enumerate}
	\item{Given the limited exploration of ST in RS UDA segmentation, this study focuses on the ST approach for UDA.}
	\item{Due to the large domain gap of cross-domain RS image, the existing ST methods cannot learn the features of the target domain well. Therefore This paper presents the first application of contrast learning to an UDA task. }
	\item{A novel loss function, based on contrastive learning, is proposed that incorporates contrast learning loss to enhance the model's learning effectiveness.}
\end{enumerate}

\section{Related Work}
\subsection{Unsupervised cross-domain adaptation for semantic segmentation}
Adversarial learning is a prevalent approach among various effective methods. Tsai et al. \cite{tsai2018} argued that there is a high degree of similarity between the source and target domains in terms of semantic layout, leading them to construct a multi-level adversarial network that exploits structural consistency in the cross-domain output space. Conversely, Vu et al. aimed to minimize the difference between the entropy distributions of the source and target domains by introducing a discriminator \cite{vu2019}. Cai et al. proposed a bidirectional adversarial learning framework to maintain semantic consistency in the segmentation of remote sensing images \cite{cai2022}.


Another typical non-adversarial unsupervised domain adaptation paradigm is self-training (ST), which has gained significant attention in cross-domain semantic segmentation in recent years. Zou et al. \cite{zou2018} first introduced a ST method for unsupervised domain-adaptive semantic segmentation. Tranheden et al. \cite{tranheden2021}, Zhou et al. \cite{zhou2022} , Hoyer et al. \cite{hoyer2022}, and Chen et al \cite{chen2022}. enhanced domain adaptation by generating trustworthy, consistent, and category-balanced pseudo-label.

\subsection{Contrastive Learning}
However, as shown in Fig. \ref{fig1}, compared with natural images in cross-domain RS images the domain gap is larger. Directly applying the ST method to cross-domain RS image semantic segmentation does not capture the feature information of the target domain image well, which leads to the performance degradation of the ST method in the target domain. Therefore, this paper tries to use contrast learning to make up for this defect.

The core principle of contrastive learning is to generate pairs of images (view pairs or positive sample pairs) that share the same potential significance \cite{hadsell2006}. The optimization objective of contrastive learning is to encourage the model to learn similar embeddings for positive sample pairs while effectively distinguishing irrelevant sample pairs (negative sample pairs). This approach has gained prominence in unsupervised self-training representation learning \cite{wu2018,oord2018,hjelm2018}. The concept of simple and effective contrastive learning was further advanced through the introduction of the Siamese network \cite{bromley1993,bachman2019,he2020,chen2020,chen2020a,chen2021a}.

In practice, the performance of contrastive learning methods is significantly enhanced, partly due to the utilization of a large number of negative samples, which can be stored in a memory bank \cite{wu2018}. For instance, the MoCo method \cite{he2020} maintains a queue of negative samples and employs a momentum encoder to improve the consistency of this queue. In contrast, the SimCLR method directly utilizes negative samples present in the current batch, though it typically requires a larger batch size to function effectively. The SimSiam method \cite{chen2021a} achieves effective feature learning by simplifying the design. Unlike other contrastive learning methods, SimSiam does not rely on negative samples but instead builds “pairs of positive samples” for training. 
In the context of remote sensing (RS) images, the richer image features and larger domain gap can lead to insufficient feature learning when using ST method. This limitation may hinder the effective learning of image features, adversely affecting the segmentation performance of the model. Therefore, enhancing the model's representation learning ability through contrastive learning is essential, as it not only improves the accuracy of feature learning but also enhances the model's performance in complex scenes.

\section{Method}
\subsection{Preliminary}
In Unsupervised domain adaptation (UDA) for remote sensing (RS), we define two sets of images collected from different satellites or locations as distinct domains. To simplify the problem, we assume that the source domain and target domain images share the same pixel resolution, denoted as $H\times W \times 3$.
Additionally, the two domains maintain consistency in the number of classes.

Let $x^{(i)}_S$ be the image and $y^{(i)}_S$ its corresponding label, with the source domain defined as
$D_S=\left \{ (x_S^{(i)},y_S^{(i)} ) \mid x_S^{(i)} \in \mathbb{R}^{H \times W \times 3} ,y_S^{(i)} \in \mathbb{R}^{H \times W \times C} \right \} _{i=1}^{N_S }$ 
, where $C$ is the number of classes. The target domain is defined as 
$D_T=\left \{ x_T^{(i)} \mid x_T^{(i)} \in \mathbb{R}^{H \times W \times 3}  \right \} _{i=1}^{N_T}$
,  where only the images $x_T^{(i)}$ are accessible, while the labels  $y_T^{(i)}$ remain unavailable. The subscripts
$S$ and $T$ denote the source and target domains, respectively, and $N_S$ and $N_T$ indicate the number of samples in the source and target domains. The representation of the source domain label $y_S$ at the spatial position $(h,w)$ is denoted as a length $C$ one-hot encoding, represented as $y^{(i,h,w)}_S$, where $h \in [1, \dots , H]$ and  $w \in [1, \dots , W]$.

If we solely rely on cross-entropy loss in the source domain for training the network $g_\theta$, it can be expressed as follows:

\begin{equation}
	\label{eq:loss_ce}
	L^{(i)}_S=- \sum_{h=1}^{H} \sum_{w=1}^{W} \sum_{c=1}^{C} y^{(i,h,w,c)}_S\cdot log(g_\theta (x^{(i)}_S)^{(h,w,c)}).
\end{equation}
In Equation (\ref{eq:loss_ce}), $g_\theta (x^{(i)}_S)$ represents the predicted outcomes for each pixel in the source domain image $x^{(i)}_S$.
However, due to the domain gap, relying solely on the source domain for training typically results in poor performance on the target domain, as the network struggles to generalize to target domain samples.

\subsection{Self-Training for UDA}

To transfer knowledge from the source domain to the target domain, the ST method employs a teacher network $t_\theta$ to generate corresponding pseudo-label for the target domain images. Mathematically, this is expressed in Equation (\ref{eq:pseudo_gen}):

\begin{equation}
	\label{eq:pseudo_gen}
	p_T^{(i,h,w,c)}=\left [ c=arg\max_{c'}t_\theta (x^{(i)}_T)^{(h,w,c)}  \right ].
\end{equation}
where $[\cdot]$ denotes the Iverson bracket. Here, $t_\theta(x_T^{(i)})$ indicates the class predictions for each pixel in the target domain image $x^{(i)}_T$.  It is important to note that gradients are not backpropagated through the teacher network. Since we cannot ascertain the correctness of the generated pseudo-label, it is necessary to evaluate the quality or confidence of the predictions for the pixels in the pseudo-label. Only those pixels with class confidence exceeding a threshold $\tau$ will be used for training. The mathematical formulation for assessing the quality or confidence of the pixel at position $(h,w)$ is given by equation (\ref{eq:pseudo_quality}):

\begin{equation}
	\label{eq:pseudo_quality}
	q_T^{(i)}=\frac{ {\textstyle \sum_{h=1}^{H}} {\textstyle \sum_{w=1}^{W}} [\max (t_\theta(x_T^{(i)})^{(h,w)})] > \tau}{H \cdot W }.
\end{equation}

In addition to training on labeled data in the source domain using Equation (\ref{eq:loss_ce}), the pseudo-label $p_T^{(i)}$ and their corresponding quality estimates $q_T^{(i)}$ from the target domain will also be incorporated into the training of the student network $g_\theta$. The loss for training in the target domain can be mathematically represented as Equation (\ref{eq:loss_tgt}):
\begin{equation}
	\label{eq:loss_tgt}
L^{(i)}_T=- \sum_{h=1}^{H} \sum_{w=1}^{W} \sum_{c=1}^{C}q_T^{(i)}\cdot p^{(i,h,w,c)}_T\cdot log(g_\theta (x^{(i)}_T)^{(h,w,c)}).
\end{equation}

During the training process, the weights of the teacher network are updated after each training iteration using the exponentially moving average (EMA) method \cite{tarvainen2017}, thereby enhancing the stability of pseudo-label generation. This can be mathematically expressed as Equation (\ref{eq:ema_update}):
\begin{equation}
	\label{eq:ema_update}
		S_{t+1}(t_\theta ) \gets \alpha  \cdot S_{t}(t_\theta) + (1-\alpha )\cdot S_{t}(g_\theta).
\end{equation}
where $S_t(\cdot)$ denotes the weights of the model at training step $t$, and the hyperparameter $\alpha \in[0,1]$ indicates the importance of the current state $S(t_\theta)$. 


\subsection{Contrastive Learning}
In the context of RS images, the richness of image features can exacerbate domain gaps and lead to insufficient feature learning when applying the ST method. These limitations may hinder effective feature learning, ultimately degrading the model's segmentation performance.
To overcome the limitations of ST methods in RS, we introduce a contrastive learning to improve methods performance in the target domain. The contrastive learning module generates two distinct views $x_T^{(i)}$ through two random augmentations. These views are processed by an encoder network $E$ , consisting of a backbone $f$ (e.g., MIT \cite{xie2021}) and a Multi-Layer Perceptron (MLP) projection head \cite{chen2020}. The weights of the encoder network $E$  remain the same while processing both views. We denote the MLP prediction head as $h$; the output of one view through the encoder network $E$ is transformed to match the representation of the other view. The outputs of the MLP prediction head and the MLP projection head can be expressed as $p_1^{(i)}=h(E(x_{T_1}^{(i)}))$ and $z_1^{(i)}=E(x_{T_2}^{(i)})$. Our objective is to optimize the negative cosine similarity between these two vectors, mathematically represented as Equation (\ref{eq:sim}):
\begin{equation}
	\label{eq:sim}
	D(p_1^{(i)}, z_2^{(i)})=-\frac{ p_1^{(i)} }{ \left \| p_1^{(i)} \right \|_2  } \cdot \frac{ z_2^{(i)} }{ \left \| z_2^{(i)} \right \|_2  }.
\end{equation}
where $ \left \| \cdot \right \|_2$ denotes $l_2$ normalization. Thus, we define the contrastive learning loss, in Equation (\ref{eq:contrast_loss}), as follows:

\begin{equation}
	\label{eq:contrast_loss}
	L_{CLR}^{(i)}=\frac{1}{2} \cdot D(p_1^{(i)}, sg( z_2^{(i)})) + \frac{1}{2} \cdot D(p_2^{(i)}, sg( z_1^{(i)})).
\end{equation}

In the loss function, the stop-gradient operation is a crucial step, treating the variables in $sg(\cdot)$ as constants. This operation effectively prevents representation collapse \cite{chen2021a}. Specifically, in the first term of Equation (\ref{eq:contrast_loss}), the encoder processing $x_{T_2}^{(i)}$ does not receive gradients from $z_2^{(i)}$, while in the second term, the encoder processing $x_{T_1}^{(i)}$ does receive gradients from $p_2^{(i)}$. This design effectively enhances the model's representation learning capabilities, improving its performance in the target domain. 

\subsection{Proposed UDA Losses}
In a given training step, we acquire the $m$-th image and its corresponding label from the source domain, represented as $x_S^{(m)}$ and $y_S^{(m)}$. Simultaneously, we obtain the $m$-th image from the target domain and its pseudo-label generated by the teacher network $t_\theta$,  denoted as $x_T^{(n)}$ and  $p_T^{(n)}$. The total loss function we mathematically define is as follows:
\begin{equation}
	\label{eq:total_loss}
	L_{total}^{(m;n)}=L_S^{(m)}+\beta \cdot L_T^{(n)} + \gamma \cdot L_{CLR}^{(n)}.
\end{equation}
where $\beta$ and $\gamma$ are balancing factors used to weigh the different losses. Through Equation (\ref{eq:total_loss}), we can comprehensively consider the losses from both the source and target domains, facilitating more effective model training.

Specifically, the loss $L_S^{(m)}$ guides the model's performance on source domain data, while $L_T^{(n)}$ 
leverages pseudo-label to drive learning in the target domain, further enhancing the model's generalization ability in this domain. Finally, $L_{CLR}^{(n)}$ introduces richer feature representations through contrastive learning, enabling the model to better adapt to potential differences between the source and target domains. By integrating these losses, we can effectively reduce the domain gap between the source and target domains, thereby improving the model's performance on the unlabeled target domain data.

\subsection{SiamSeg Network Architecture}
\subsubsection{overall}

\begin{figure*}[!t]
	\centering
	\begin{minipage}{0.49\linewidth}
			\centering
		\includegraphics[width=1.0\linewidth]{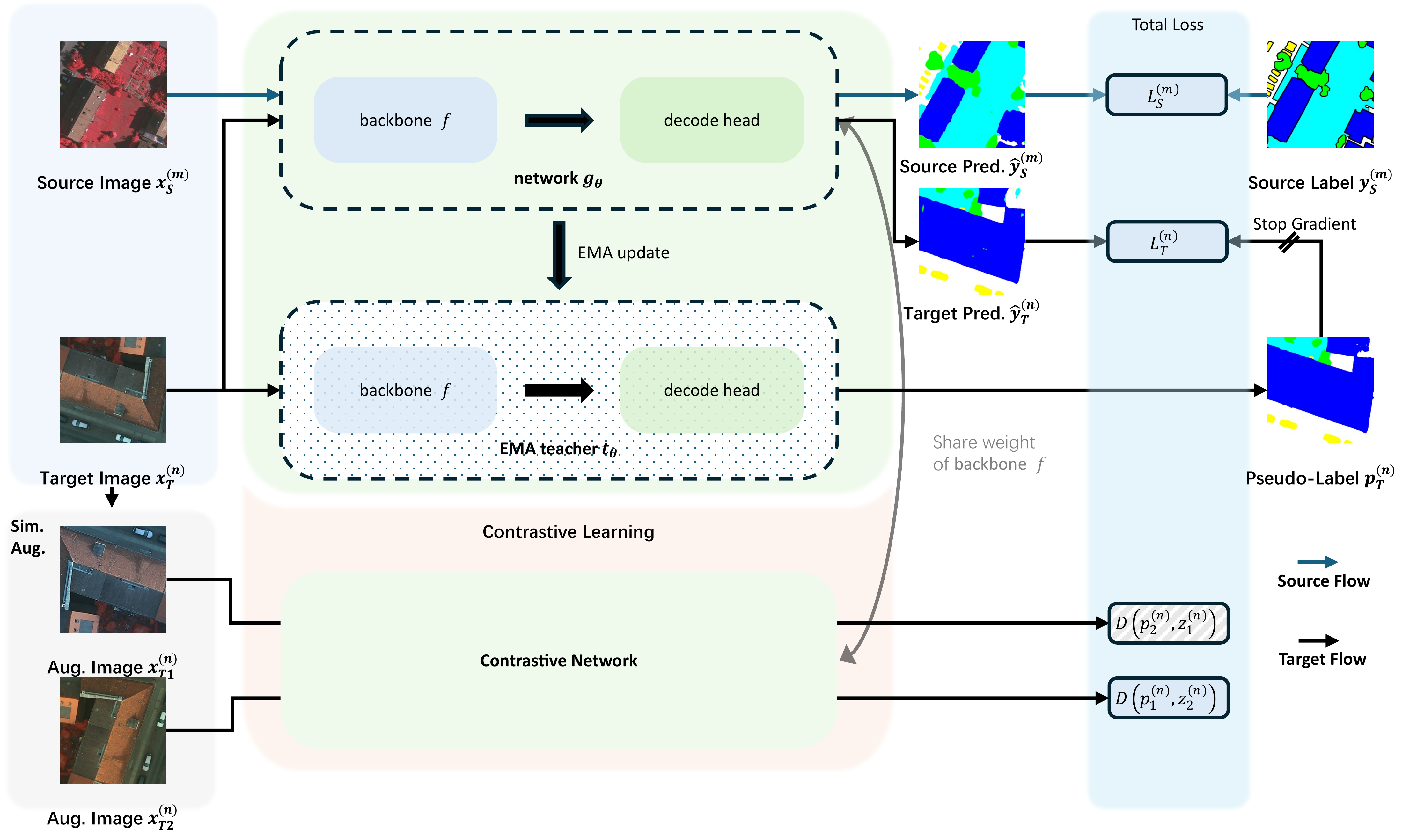}
		\caption{
			Overall of SiamSeg. The network \( g_\theta \) is designed for image segmentation and comprises a feature extraction backbone \( f \) and a decoding head, an EMA teacher network \( t_\theta \) and a contrastive network.
		}
		\label{fig:network_architecture}
	\end{minipage}
	\begin{minipage}{0.49\linewidth}
		\centering
		\includegraphics[width=1.0\linewidth]{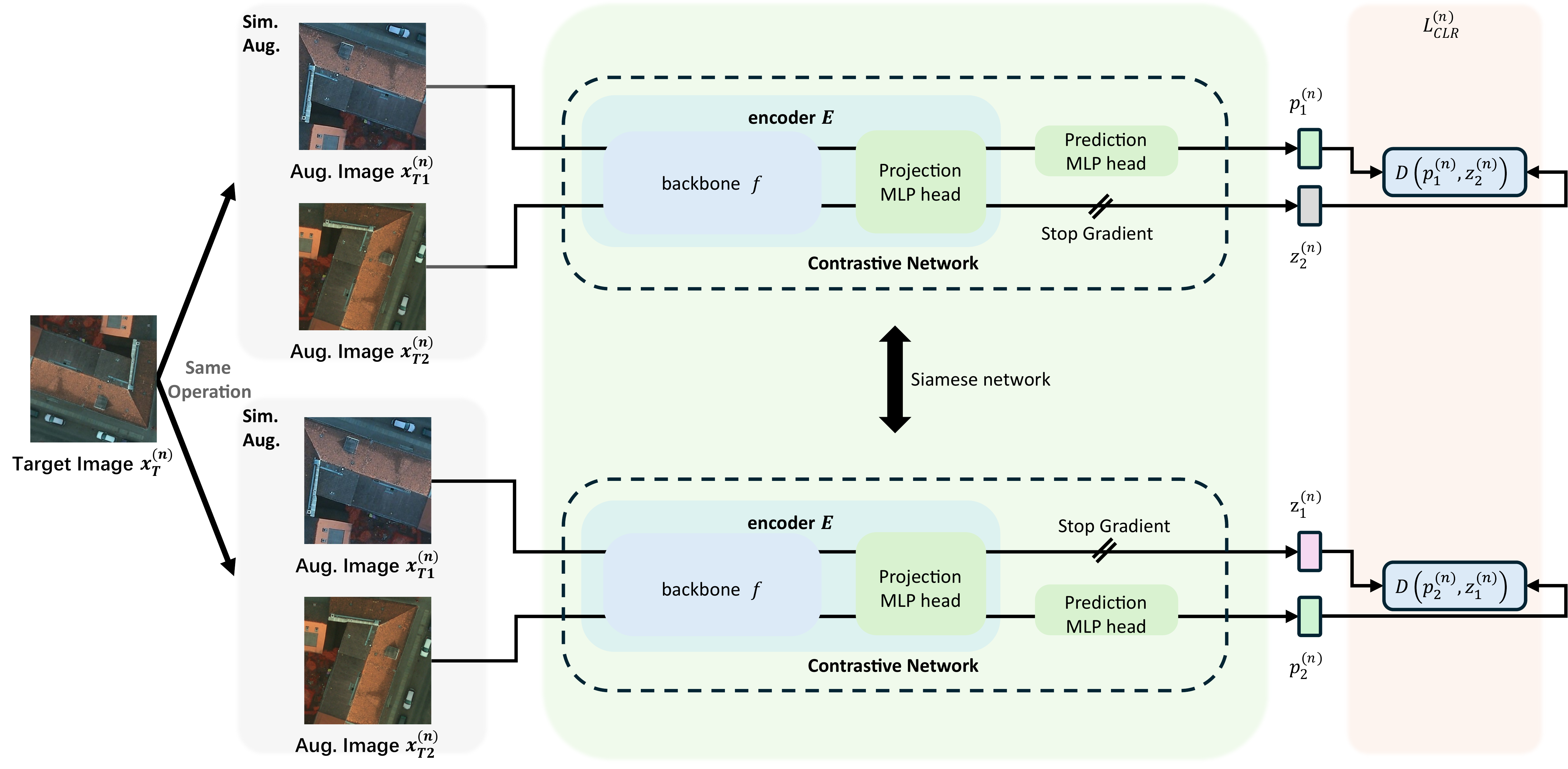}
		\caption{Detail of Contrastive Network. This figure illustrates the architecture of the Siamese network used for contrastive learning. The network consists of two identical sub-networks that share the same model weights, ensuring consistency in feature extraction.}
		\label{fig:contrast_network}
	\end{minipage}
\end{figure*}
As illustrated in Figure \ref{fig:network_architecture}, the network $g_\theta$ is a model designed for image segmentation tasks. Its architecture comprises a feature extraction backbone $f$  and a decoding head responsible for segmentation predictions. To enhance the model's stability and performance, we introduce an EMA teacher network $t_\theta$, which has an identical architecture to $g_\theta$ but does not backpropagate gradients during training. 

In the contrastive learning module, the network structure similarly includes the backbone $f$, a projection Multi-Layer Perceptron (MLP) head, and a prediction MLP head. Notably, the backbone $f$ shares weights with the backbone of the segmentation network $g_\theta$ to enhance the consistency and effectiveness of feature representations.

During training, as indicated by the blue arrows in the figure, the source image \( x_S^{(m)} \) and its corresponding source label \( y_S^{(m)} \) are utilized to initially train the network \( g_\theta \) through supervised learning. In the step denoted by the black arrows, the target image \( x_T^{(n)} \) is processed by the EMA teacher network \( t_\theta \), generating pseudo-label \( p_T^{(n)} \) to replace the inaccessible target labels \( y_T^{(n)} \) during training. 
Furthermore, by applying data augmentation to the target image \( x_T^{(n)} \), we generate two distinct views, which together form positive sample pairs and provide rich training signals, with contrastive learning, for model. 

\subsubsection{detail of contrastive learning} \label{detail_of_cl}
 

The implementation of contrastive learning utilizes the Siamese network architecture, a simple yet effective strategy \cite{bromley1993,he2020,chen2021a}. As shown in Fig. \ref{fig:contrast_network}, the contrastive learning network comprises two identical sub-networks that share the same model weights, ensuring consistency during feature extraction and facilitating effective contrastive learning. Specifically, an input image \( x_T^{(n)} \) undergoes data augmentation (eg., Resize Crop, Color Jitter, Gray Scale, Gaussian Blur, Filp), generating two distinct views, \( x_{T1}^{(n)} \) and \( x_{T2}^{(n)} \). These views are processed by the contrastive networks. Despite sharing parameters, their outputs differ slightly. In the upper workflow, the view \( x_{T1}^{(n)} \) is first encoded by the encoder \( E \) and then processed through the prediction MLP head \( h \) to yield output \( \mathbf{p}_1^{(n)} \). In contrast, the view \( x_{T2}^{(n)} \) is also passed through the encoder to produce output \( \mathbf{z}_2^{(n)} \) but does not go through the prediction MLP head; instead, it employs a stop-gradient operation. To maximize the similarity between outputs \( \mathbf{p}_1^{(n)} \) and \( \mathbf{z}_2^{(n)} \), we optimize using Equation \ref{eq:contrast_loss}. This process encourages the model to fully exploit the similarities between positive sample pairs, thereby enhancing the effectiveness and robustness of feature representation. The lower workflow mirrors the upper one, ensuring the unity and consistency of the overall contrastive learning process. Through this structured design, the Siamese network effectively performs feature learning, providing strong representational capabilities for our RS tasks.

\section{EXPERIMENTAL RESULTS AND ANALYSIS}

\subsection{Datasets} \label{dataset}
To evaluate the proposed method's performance in cross-domain remote sensing (RS) image segmentation tasks, we selected three benchmark datasets: Potsdam, Vaihingen, and LoveDA. 
\subsubsection{Potsdam and Vaihingen datasets}
The Potsdam and Vaihingen datasets are part of the ISPRS 2D semantic segmentation benchmarks \cite{rottensteiner2014}. The Potsdam (POT) dataset comprises $38$ remote sensing images with a resolution of $6000 \times 6000$ pixels and a ground sampling distance of $5$ meters. It features three different image modalities: IRRG, RGB, and RGBIR, where the first two modalities have three channels, and the last one has four. In this study, we primarily utilize the first two modalities. The Vaihingen (VAI) dataset contains $33$ remote sensing images with resolutions ranging from $1996 \times 1996$ to $3816 \times 2550$ pixels, with a ground sampling distance of $9$ centimeters. This dataset includes only one image modality (IRRG). Both datasets share six common classes: impervious surfaces, buildings, low vegetation, trees, cars, and clutter/background.

To reduce computational overhead, we cropped the images to a smaller size of $512\times512$ pixels. For the POT and VAI datasets, we used cropping strides of $512$ and $256$, resulting in $4598$ and $1696$ images, respectively. Subsequently, we split the POT and VAI datasets into training and testing sets, following previous wokrs \cite{li2022,zhao2024self}. In the POT dataset, the number of images in the training and testing sets is $2904$ and $1694$, respectively, while in the VAI dataset, these numbers are $1296$ and $440$. We established four cross-domain remote sensing semantic segmentation tasks:


\begin{itemize}
	\item{Potsdam IR-R-G to Vaihingen IR-R-G (POT IRRG → VAI IRRG).}
	\item{Vaihingen IR-R-G to Potsdam IR-R-G (VAI IRRG → POT IRRG).}
	\item{Potsdam R-G-B to Vaihingen IR-R-G (POT RGB → VAI IRRG).}
	\item{Vaihingen IR-R-G to Potsdam R-G-B (VAI IRRG → POT RGB).}
\end{itemize}

\subsubsection{LoveDA dataset}
The LoveDA dataset was recently proposed to address semantic segmentation and domain adaptation challenges in remote sensing. It consists of $5987$ high-resolution ($1024\times1024$) RS images sourced from Nanjing, Guangzhou, and Wuhan \cite{wang2021}. The LoveDA dataset contains two distinct domains: urban and rural, aimed at challenging the model's generalization capability between different geographical elements.

Within the dataset, there are $1883$ urban images, which are further divided into $1156$ training samples and $677$ validation samples as mentioned in \cite{wang2021}. In the rural domain, there are $2358$ images, with $1366$ for training and $992$ for validation. For the LoveDA dataset, we designed a cross-domain remote sensing semantic segmentation task:

\begin{itemize}
	\item{Rural to Urban (Rural → Urban).}
\end{itemize}

Through these five different tasks, we aim to assess the proposed method's adaptability and performance across various geographical environments, providing new perspectives and data support for cross-domain semantic segmentation research.

\subsection{Evaluation Metrics}
In this study, we adopt the F1-score (F1) and Intersection over Union (IoU) as evaluation metrics, following previous methods in the field of remote sensing (RS) semantic segmentation domain adaptation (UDA). The specific calculation formulas are as follows:
\begin{equation}
	\label{eq:iou}
	\text{IoU} = \frac{TP}{TP + FP + FN}.
\end{equation}

\begin{equation}
	\label{eq:f1}
	\text{F1} = \frac{2 \times TP}{2TP + FP + FN}.
\end{equation}

In equations (\ref{eq:iou}) and (\ref{eq:f1}), \(TP\) represents true positives, \(FP\) denotes false positives, and \(FN\) indicates false negatives. The IoU, also known as the Jaccard index, and the F1-score, referred to as the Dice coefficient, effectively reflect the accuracy and reliability of model performance in semantic segmentation tasks. 

\subsection{Implementation Details}
\subsubsection{Augmentation}

This study is implemented using the PyTorch framework \cite{paszke2019} and MMSegmentation \cite{contributors2023}. We utilize the data preprocessing pipeline provided by MMSegmentation, which includes operations such as random image resizing, cropping, and flipping. In addition, inspired by the DACS approach \cite{tranheden2021}, we incorporate color jitter, Gaussian blur, and ClassMix \cite{olsson2021} to enhance the dataset and improve feature robustness. During the contrastive learning stage, we generate two different views of the image by applying augmentations such as cropping, color jitter, Gaussian blur, grayscale, and flipping.

\subsubsection{Network Architecture Details}

Considering the outstanding performance of the transformer-based Segformer \cite{xie2021} in semantic segmentation tasks, we chose Mix Vision Transformers (MiT) \cite{xie2021} as the backbone network \(f\). This model was pre-trained on ImageNet \cite{deng2009}. In semantic segmentation, capturing both global and local features is critical, and feature fusion strategies are often used to improve segmentation performance. For this reason, we adopt the context-aware multi-scale feature fusion decoder head designed by Hoyer et al. \cite{hoyer2022}, given its superior performance in UDA-based semantic segmentation. 
\subsubsection{Training}
During training, the AdamW optimizer is applied to network \(g_\theta\), with hyperparameters set as betas=(0.9, 0.999) and a weight decay of 0.01. The learning rate for the backbone \(f\) is set to \(6 \times 10^{-4}\), while the decoder head, projection MLP head, and prediction MLP head have learning rates of \(6 \times 10^{-5}\). For learning rate scheduling, a linear warm-up strategy is employed for the first 1,500 iterations, followed by linear decay with a decay rate of 0.01. The decay coefficient \(\alpha\) for the exponential moving average teacher network \(t_\theta\) is set to 0.99.

In the pseudo-label generation phase, the temperature parameter \(\tau\) in Equation \ref{eq:pseudo_gen} is set to 0.999. The total number of training iterations is 40,000, with a batch size of 12, containing 6 source domain and 6 target domain images. 

During the contrastive learning process, data augmentations (Sim. Aug.) use a crop size of \(size=(512,512)\) and scale range \(scale=(0.6,1.0)\). Color jitter parameters are set to a brightness, contrast, saturation, and hue of 0.25 each, with a random application probability of 0.6. Grayscale has a probability of 0.2, Gaussian blur has a probability of 0.5, and both horizontal and vertical flips are applied with a probability of 0.5. These settings aim to enhance model generalization through data augmentation, improving performance in cross-domain remote sensing image segmentation tasks.


All experiments are conducted within the MMSegmentation framework to ensure consistency and reproducibility. Moreover, all model training is performed using Nvidia A100 GPU$\times 4$, enhancing training efficiency and performance. 

\subsection{Experimental Results}
\subsubsection{Quantitative Results}

\paragraph{Cross-domain RS Image Semantic Segmentation on POT and VAI}

For the Potsdam (POT) and Vaihingen (VAI) datasets, as described in Section \ref{dataset}, we established four sets of cross-domain remote sensing (RS) semantic segmentation tasks. In this subsection, we validate the effectiveness of the proposed SimSeg method through a series of comprehensive experimental results.

Since the backbone network used in this study is Mix Transformers (MiT) \cite{xie2021}, Segformer is selected as the baseline for comparison. The Segformer model was trained solely on the source domain and tested directly on the target domain. Additionally, we evaluate multiple comparison methods, including AdaptSegNet \cite{tsai2018}, ProDA \cite{zhang2021}, and several RS-specific segmentation methods, such as DualGAN \cite{li2021}, CIA-UDA \cite{ni2023}, and DNT \cite{chen2022a}. These comparisons allow us to demonstrate not only the superior performance of SimSeg in cross-domain RS image semantic segmentation but also to provide valuable insights for further research in domain adaptation.

From Table \ref{tab:POT_IRRG_2_VAI_IRRG}, \ref{tab:POT_RGB_2_VAI_IRRG}, \ref{tab:VAI_IRRG_2_POT_IRRG} and \ref{tab:VAI_IRRG_2_POT_RGB}, we can find that for some of the methods that use Deeplab as a backbone, such as AdaptSegNet, ProDA and DualGAN, the IoU is lower than the prediction accuracies obtained using Segformer, which is trained only on the source domain, in most cases. This is due to the fact that Segformer is a transformer-based method, while DeepLab is a convolution-based method. For the more complex image features of RS images, the attention mechanism of transformer is able to better utilize the feature context of the image, so some of the methods still have lower accuracy when facing RS images, even for UDA methods. As for CIA-UDA and DNT, which are methods designed for the characteristics of RS images, they achieve competitive performance in segmentation results. For the methods Siamseg and SiamSeg without C.L. in this paper, without the addition of contrastive learning, the reliance on ST still achieves good performance, which shows that the ST method is an effective UDA method. However, since the domain shift of RS images varies greatly between different domains, this leads to insufficient learning of the target domain image. And after adding contrastive learning, it can be observed that the method predicts a significant increase in IoU. Thus indicating that the addition of contrastive learning does allow the model to learn more image features of the target domain, thus greatly enhancing the performance.

%
%

\begin{table*}[]
	\caption{Cross-domain RS image semantic segmentation results from Potsdam IRRG to Vaihingen IRRG. The best and second-best results are highlighted in \textbf{bold} and {\ul{underlined}}, respectively, in each column. The evaluation metrics used are IoU and F1-score, where F1-score is abbreviated as F1. All values are presented as percentages (\%), with larger values indicating better performance. The last column provides the average scores across all categories. Note that Segformer was trained only on the source domain and then tested directly on the target domain. "C.L." in the table refers to contrastive learning as proposed in this study.}
	\label{tab:POT_IRRG_2_VAI_IRRG}
	\begin{tabular}{c|cc|cc|cc|cc|cc|cc|cc}
		\hline
		&
		\multicolumn{2}{c|}{Clutter} &
		\multicolumn{2}{c|}{Car} &
		\multicolumn{2}{c|}{Tree} &
		\multicolumn{2}{c|}{Low Vegetation} &
		\multicolumn{2}{c|}{Building} &
		\multicolumn{2}{c|}{Impervious Surface} &
		\multicolumn{2}{c}{Overall} \\ \cline{2-15} 
		\multirow{-2}{*}{Method} &
		\multicolumn{1}{c|}{IoU} &
		F1 &
		\multicolumn{1}{c|}{IoU} &
		F1 &
		\multicolumn{1}{c|}{IoU} &
		F1 &
		\multicolumn{1}{c|}{IoU} &
		F1 &
		\multicolumn{1}{c|}{IoU} &
		F1 &
		\multicolumn{1}{c|}{IoU} &
		F1 &
		\multicolumn{1}{c|}{mIoU} &
		mF1 \\ \hline
		Segformer \cite{xie2021} (src.) &
		\multicolumn{1}{c|}{4.22} &
		9.47 &
		\multicolumn{1}{c|}{31.13} &
		47.89 &
		\multicolumn{1}{c|}{66.31} &
		78.87 &
		\multicolumn{1}{c|}{44.47} &
		60.38 &
		\multicolumn{1}{c|}{75.5} &
		87.95 &
		\multicolumn{1}{c|}{61.03} &
		76.07 &
		\multicolumn{1}{c|}{46.11} &
		60.11 \\
		\rowcolor[HTML]{EFEFEF} 
		AdaptSegNet \cite{tsai2018} &
		\multicolumn{1}{c|}{\cellcolor[HTML]{EFEFEF}4.6} &
		8.76 &
		\multicolumn{1}{c|}{\cellcolor[HTML]{EFEFEF}6.4} &
		11.99 &
		\multicolumn{1}{c|}{\cellcolor[HTML]{EFEFEF}52.65} &
		68.96 &
		\multicolumn{1}{c|}{\cellcolor[HTML]{EFEFEF}28.98} &
		44.91 &
		\multicolumn{1}{c|}{\cellcolor[HTML]{EFEFEF}63.14} &
		77.4 &
		\multicolumn{1}{c|}{\cellcolor[HTML]{EFEFEF}54.39} &
		70.39 &
		\multicolumn{1}{c|}{\cellcolor[HTML]{EFEFEF}35.02} &
		47.05 \\
		ProDA \cite{zhang2021} &
		\multicolumn{1}{c|}{3.99} &
		8.21 &
		\multicolumn{1}{c|}{39.2} &
		56.52 &
		\multicolumn{1}{c|}{56.26} &
		72.09 &
		\multicolumn{1}{c|}{34.49} &
		51.65 &
		\multicolumn{1}{c|}{71.61} &
		82.95 &
		\multicolumn{1}{c|}{65.51} &
		76.85 &
		\multicolumn{1}{c|}{44.68} &
		58.05 \\
		\rowcolor[HTML]{EFEFEF} 
		DualGAN \cite{li2021} &
		\multicolumn{1}{c|}{\cellcolor[HTML]{EFEFEF}\textbf{29.66}} &
		\textbf{45.65} &
		\multicolumn{1}{c|}{\cellcolor[HTML]{EFEFEF}34.34} &
		51.09 &
		\multicolumn{1}{c|}{\cellcolor[HTML]{EFEFEF}57.66} &
		73.14 &
		\multicolumn{1}{c|}{\cellcolor[HTML]{EFEFEF}38.87} &
		55.97 &
		\multicolumn{1}{c|}{\cellcolor[HTML]{EFEFEF}62.3} &
		76.77 &
		\multicolumn{1}{c|}{\cellcolor[HTML]{EFEFEF}49.41} &
		66.13 &
		\multicolumn{1}{c|}{\cellcolor[HTML]{EFEFEF}45.38} &
		61.43 \\
		CIA-UDA\cite{ni2023} &
		\multicolumn{1}{c|}{27.8} &
		43.51 &
		\multicolumn{1}{c|}{52.91} &
		69.21 &
		\multicolumn{1}{c|}{64.11} &
		78.13 &
		\multicolumn{1}{c|}{48.03} &
		64.9 &
		\multicolumn{1}{c|}{75.13} &
		85.8 &
		\multicolumn{1}{c|}{63.28} &
		77.51 &
		\multicolumn{1}{c|}{55.21} &
		69.84 \\
		\rowcolor[HTML]{EFEFEF} 
		DNT\cite{chen2022a} &
		\multicolumn{1}{c|}{\cellcolor[HTML]{EFEFEF}14.77} &
		25.74 &
		\multicolumn{1}{c|}{\cellcolor[HTML]{EFEFEF}53.88} &
		70.03 &
		\multicolumn{1}{c|}{\cellcolor[HTML]{EFEFEF}59.19} &
		74.37 &
		\multicolumn{1}{c|}{\cellcolor[HTML]{EFEFEF}47.51} &
		64.42 &
		\multicolumn{1}{c|}{\cellcolor[HTML]{EFEFEF}80.04} &
		88.91 &
		\multicolumn{1}{c|}{\cellcolor[HTML]{EFEFEF}69.74} &
		82.18 &
		\multicolumn{1}{c|}{\cellcolor[HTML]{EFEFEF}54.19} &
		57.61 \\
		SiamSeg w./o. C.L. &
		\multicolumn{1}{c|}{18.14} &
		30.71 &
		\multicolumn{1}{c|}{\textbf{56.57}} &
		\textbf{72.26} &
		\multicolumn{1}{c|}{{\ul 73.95}} &
		{\ul 85.03} &
		\multicolumn{1}{c|}{{\ul 62.26}} &
		{\ul 76.74} &
		\multicolumn{1}{c|}{{\ul 87.47}} &
		{\ul 93.32} &
		\multicolumn{1}{c|}{{\ul 79.74}} &
		{\ul 88.73} &
		\multicolumn{1}{c|}{{\ul 63.02}} &
		{\ul 74.46} \\
		\rowcolor[HTML]{EFEFEF} 
		SiamSeg &
		\multicolumn{1}{c|}{\cellcolor[HTML]{EFEFEF}{\ul 27.6}} &
		{\ul 43.26} &
		\multicolumn{1}{c|}{\cellcolor[HTML]{EFEFEF}{\ul 52.51}} &
		{\ul 68.86} &
		\multicolumn{1}{c|}{\cellcolor[HTML]{EFEFEF}\textbf{77.69}} &
		\textbf{87.44} &
		\multicolumn{1}{c|}{\cellcolor[HTML]{EFEFEF}\textbf{65.73}} &
		\textbf{79.32} &
		\multicolumn{1}{c|}{\cellcolor[HTML]{EFEFEF}\textbf{89}} &
		\textbf{94.18} &
		\multicolumn{1}{c|}{\cellcolor[HTML]{EFEFEF}\textbf{80.75}} &
		\textbf{89.35} &
		\multicolumn{1}{c|}{\cellcolor[HTML]{EFEFEF}\textbf{65.55}} &
		\textbf{77.07}
	\end{tabular}
\end{table*}

\begin{table*}[]
	\caption{Cross-domain RS image semantic segmentation results from Potsdam RGB to Vaihingen IRRG.}
	\label{tab:POT_RGB_2_VAI_IRRG}
	\begin{tabular}{c|cc|cc|cc|cc|cc|cc|cc}
		\hline
		&
		\multicolumn{2}{c|}{Clutter} &
		\multicolumn{2}{c|}{Car} &
		\multicolumn{2}{c|}{Tree} &
		\multicolumn{2}{c|}{Low Vegetation} &
		\multicolumn{2}{c|}{Building} &
		\multicolumn{2}{c|}{Impervious Surface} &
		\multicolumn{2}{c}{Overall} \\ \cline{2-15} 
		\multirow{-2}{*}{Method} &
		\multicolumn{1}{c|}{IoU} &
		F1 &
		\multicolumn{1}{c|}{IoU} &
		F1 &
		\multicolumn{1}{c|}{IoU} &
		F1 &
		\multicolumn{1}{c|}{IoU} &
		F1 &
		\multicolumn{1}{c|}{IoU} &
		F1 &
		\multicolumn{1}{c|}{IoU} &
		F1 &
		\multicolumn{1}{c|}{mIoU} &
		mF1 \\ \hline
		Segformer \cite{xie2021} (src.) &
		\multicolumn{1}{c|}{1.43} &
		2.81 &
		\multicolumn{1}{c|}{37.97} &
		55.04 &
		\multicolumn{1}{c|}{52.62} &
		68.96 &
		\multicolumn{1}{c|}{5.18} &
		9.85 &
		\multicolumn{1}{c|}{73.18} &
		84.51 &
		\multicolumn{1}{c|}{51.34} &
		67.85 &
		\multicolumn{1}{c|}{36.95} &
		48.17 \\
		\rowcolor[HTML]{EFEFEF} 
		AdaptSegNet \cite{tsai2018} &
		\multicolumn{1}{c|}{\cellcolor[HTML]{EFEFEF}2.29} &
		5.81 &
		\multicolumn{1}{c|}{\cellcolor[HTML]{EFEFEF}10.25} &
		18.45 &
		\multicolumn{1}{c|}{\cellcolor[HTML]{EFEFEF}55.51} &
		68.02 &
		\multicolumn{1}{c|}{\cellcolor[HTML]{EFEFEF}12.75} &
		22.61 &
		\multicolumn{1}{c|}{\cellcolor[HTML]{EFEFEF}60.72} &
		75.55 &
		\multicolumn{1}{c|}{\cellcolor[HTML]{EFEFEF}51.26} &
		67.77 &
		\multicolumn{1}{c|}{\cellcolor[HTML]{EFEFEF}31.58} &
		43.05 \\
		ProDA \cite{zhang2021}&
		\multicolumn{1}{c|}{2.39} &
		5.09 &
		\multicolumn{1}{c|}{31.56} &
		48.16 &
		\multicolumn{1}{c|}{49.11} &
		65.86 &
		\multicolumn{1}{c|}{32.44} &
		49.06 &
		\multicolumn{1}{c|}{68.94} &
		81.89 &
		\multicolumn{1}{c|}{49.04} &
		66.11 &
		\multicolumn{1}{c|}{38.91} &
		52.7 \\
		\rowcolor[HTML]{EFEFEF} 
		DualGAN \cite{li2021} &
		\multicolumn{1}{c|}{\cellcolor[HTML]{EFEFEF}3.94} &
		13.88 &
		\multicolumn{1}{c|}{\cellcolor[HTML]{EFEFEF}40.31} &
		57.88 &
		\multicolumn{1}{c|}{\cellcolor[HTML]{EFEFEF}55.82} &
		70.61 &
		\multicolumn{1}{c|}{\cellcolor[HTML]{EFEFEF}27.85} &
		42.17 &
		\multicolumn{1}{c|}{\cellcolor[HTML]{EFEFEF}65.44} &
		83 &
		\multicolumn{1}{c|}{\cellcolor[HTML]{EFEFEF}49.16} &
		61.33 &
		\multicolumn{1}{c|}{\cellcolor[HTML]{EFEFEF}39.93} &
		54.82 \\
		CIA-UDA \cite{ni2023}&
		\multicolumn{1}{c|}{\textbf{13.5}} &
		\textbf{23.78} &
		\multicolumn{1}{c|}{\textbf{55.58}} &
		{\ul 68.66} &
		\multicolumn{1}{c|}{63.43} &
		77.62 &
		\multicolumn{1}{c|}{33.31} &
		49.97 &
		\multicolumn{1}{c|}{79.71} &
		88.71 &
		\multicolumn{1}{c|}{62.63} &
		77.02 &
		\multicolumn{1}{c|}{50.81} &
		64.29 \\
		\rowcolor[HTML]{EFEFEF} 
		DNT \cite{chen2022a} &
		\multicolumn{1}{c|}{\cellcolor[HTML]{EFEFEF}11.55} &
		20.71 &
		\multicolumn{1}{c|}{\cellcolor[HTML]{EFEFEF}{\ul 52.64}} &
		\textbf{68.97} &
		\multicolumn{1}{c|}{\cellcolor[HTML]{EFEFEF}58.43} &
		73.76 &
		\multicolumn{1}{c|}{\cellcolor[HTML]{EFEFEF}\textbf{43.63}} &
		\textbf{61.5} &
		\multicolumn{1}{c|}{\cellcolor[HTML]{EFEFEF}{\ul 81.09}} &
		{\ul 89.56} &
		\multicolumn{1}{c|}{\cellcolor[HTML]{EFEFEF}67.94} &
		80.91 &
		\multicolumn{1}{c|}{\cellcolor[HTML]{EFEFEF}{\ul 52.6}} &
		{\ul 65.83} \\
		SiamSeg w./o. C.L. &
		\multicolumn{1}{c|}{6.66} &
		12.49 &
		\multicolumn{1}{c|}{51.85} &
		68.29 &
		\multicolumn{1}{c|}{{\ul 68.06}} &
		{\ul 80.99} &
		\multicolumn{1}{c|}{28.39} &
		44.22 &
		\multicolumn{1}{c|}{83.6} &
		91.07 &
		\multicolumn{1}{c|}{{\ul 69.23}} &
		{\ul 81.82} &
		\multicolumn{1}{c|}{51.3} &
		63.15 \\
		\rowcolor[HTML]{EFEFEF} 
		SiamSeg &
		\multicolumn{1}{c|}{\cellcolor[HTML]{EFEFEF}{\ul 13.23}} &
		{\ul 23.37} &
		\multicolumn{1}{c|}{\cellcolor[HTML]{EFEFEF}51.14} &
		67.67 &
		\multicolumn{1}{c|}{\cellcolor[HTML]{EFEFEF}\textbf{71.09}} &
		\textbf{83.1} &
		\multicolumn{1}{c|}{\cellcolor[HTML]{EFEFEF}{\ul 40.1}} &
		{\ul 57.24} &
		\multicolumn{1}{c|}{\cellcolor[HTML]{EFEFEF}\textbf{81.56}} &
		\textbf{89.85} &
		\multicolumn{1}{c|}{\cellcolor[HTML]{EFEFEF}\textbf{73.15}} &
		\textbf{84.49} &
		\multicolumn{1}{c|}{\cellcolor[HTML]{EFEFEF}\textbf{55.04}} &
		\multicolumn{1}{c|}{\cellcolor[HTML]{EFEFEF}\textbf{67.62}}
	\end{tabular}
\end{table*}

\begin{table*}[]
	\caption{Cross-domain RS image semantic segmentation results from Vaihingen IRRG to Potsdam IRRG.}
	\label{tab:VAI_IRRG_2_POT_IRRG}
	\begin{tabular}{c|cc|cc|cc|cc|cc|cc|cc}
		\hline
		&
		\multicolumn{2}{c|}{Clutter} &
		\multicolumn{2}{c|}{Car} &
		\multicolumn{2}{c|}{Tree} &
		\multicolumn{2}{c|}{Low Vegetation} &
		\multicolumn{2}{c|}{Building} &
		\multicolumn{2}{c|}{Impervious Surface} &
		\multicolumn{2}{c}{Overall} \\ \cline{2-15} 
		\multirow{-2}{*}{Method} &
		\multicolumn{1}{c|}{IoU} &
		F1 &
		\multicolumn{1}{c|}{IoU} &
		F1 &
		\multicolumn{1}{c|}{IoU} &
		F1 &
		\multicolumn{1}{c|}{IoU} &
		F1 &
		\multicolumn{1}{c|}{IoU} &
		F1 &
		\multicolumn{1}{c|}{IoU} &
		F1 &
		\multicolumn{1}{c|}{mIoU} &
		mF1 \\ \hline 
		Segformer \cite{xie2021} (src.) &
		\multicolumn{1}{c|}{1.08} &
		2.56 &
		\multicolumn{1}{c|}{58.99} &
		73.14 &
		\multicolumn{1}{c|}{30.07} &
		46.24 &
		\multicolumn{1}{c|}{51.91} &
		68.82 &
		\multicolumn{1}{c|}{74.85} &
		87.18 &
		\multicolumn{1}{c|}{60.63} &
		76.47 &
		\multicolumn{1}{c|}{46.29} &
		59.08 \\ 
		\rowcolor[HTML]{EFEFEF} 
		AdaptSegNet \cite{tsai2018}&
		\multicolumn{1}{c|}{\cellcolor[HTML]{EFEFEF}8.36} &
		15.33 &
		\multicolumn{1}{c|}{\cellcolor[HTML]{EFEFEF}40.95} &
		58.11 &
		\multicolumn{1}{c|}{\cellcolor[HTML]{EFEFEF}22.59} &
		36.79 &
		\multicolumn{1}{c|}{\cellcolor[HTML]{EFEFEF}34.43} &
		64.5 &
		\multicolumn{1}{c|}{\cellcolor[HTML]{EFEFEF}48.01} &
		63.41 &
		\multicolumn{1}{c|}{\cellcolor[HTML]{EFEFEF}49.55} &
		64.64 &
		\multicolumn{1}{c|}{\cellcolor[HTML]{EFEFEF}33.98} &
		49.96 \\
		ProDA \cite{zhang2021}&
		\multicolumn{1}{c|}{10.63} &
		19.21 &
		\multicolumn{1}{c|}{46.78} &
		63.74 &
		\multicolumn{1}{c|}{31.59} &
		48.02 &
		\multicolumn{1}{c|}{40.55} &
		57.71 &
		\multicolumn{1}{c|}{56.85} &
		72.49 &
		\multicolumn{1}{c|}{44.7} &
		61.72 &
		\multicolumn{1}{c|}{38.51} &
		53.82 \\
		\rowcolor[HTML]{EFEFEF} 
		DualGAN \cite{li2021}&
		\multicolumn{1}{c|}{\cellcolor[HTML]{EFEFEF}{\ul 11.48}} &
		{\ul 20.56} &
		\multicolumn{1}{c|}{\cellcolor[HTML]{EFEFEF}48.49} &
		65.31 &
		\multicolumn{1}{c|}{\cellcolor[HTML]{EFEFEF}34.98} &
		51.82 &
		\multicolumn{1}{c|}{\cellcolor[HTML]{EFEFEF}36.5} &
		53.48 &
		\multicolumn{1}{c|}{\cellcolor[HTML]{EFEFEF}53.37} &
		69.59 &
		\multicolumn{1}{c|}{\cellcolor[HTML]{EFEFEF}51.01} &
		67.53 &
		\multicolumn{1}{c|}{\cellcolor[HTML]{EFEFEF}39.3} &
		54.71 \\
		CIA-UDA \cite{ni2023}&
		\multicolumn{1}{c|}{10.87} &
		19.61 &
		\multicolumn{1}{c|}{65.35} &
		79.04 &
		\multicolumn{1}{c|}{47.74} &
		64.63 &
		\multicolumn{1}{c|}{54.4} &
		70.47 &
		\multicolumn{1}{c|}{72.31} &
		83.93 &
		\multicolumn{1}{c|}{62.74} &
		77.11 &
		\multicolumn{1}{c|}{52.23} &
		65.8 \\
		\rowcolor[HTML]{EFEFEF} 
		DNT \cite{chen2022a}&
		\multicolumn{1}{c|}{\cellcolor[HTML]{EFEFEF}\textbf{11.51}} &
		\textbf{20.65} &
		\multicolumn{1}{c|}{\cellcolor[HTML]{EFEFEF}49.5} &
		66.22 &
		\multicolumn{1}{c|}{\cellcolor[HTML]{EFEFEF}35.46} &
		52.36 &
		\multicolumn{1}{c|}{\cellcolor[HTML]{EFEFEF}37.61} &
		54.67 &
		\multicolumn{1}{c|}{\cellcolor[HTML]{EFEFEF}66.41} &
		79.81 &
		\multicolumn{1}{c|}{\cellcolor[HTML]{EFEFEF}61.91} &
		76.48 &
		\multicolumn{1}{c|}{\cellcolor[HTML]{EFEFEF}43.74} &
		58.36 \\ 
		SiamSeg w./o. C.L. &
		\multicolumn{1}{c|}{3.04} &
		5.91 &
		\multicolumn{1}{c|}{\textbf{76.29}} &
		\textbf{86.55} &
		\multicolumn{1}{c|}{{\ul 58.51}} &
		{\ul 73.82} &
		\multicolumn{1}{c|}{{\ul 66.67}} &
		{\ul 80} &
		\multicolumn{1}{c|}{\textbf{83.7}} &
		\textbf{91.13} &
		\multicolumn{1}{c|}{{\ul 75.87}} &
		{\ul 86.28} &
		\multicolumn{1}{c|}{{\ul 60.68}} &
		{\ul 70.62} \\
		\rowcolor[HTML]{EFEFEF} 
		SiamSeg &
		\multicolumn{1}{c|}{\cellcolor[HTML]{EFEFEF}5.2} &
		9.89 &
		\multicolumn{1}{c|}{\cellcolor[HTML]{EFEFEF}{\ul 76.19}} &
		{\ul 86.48} &
		\multicolumn{1}{c|}{\cellcolor[HTML]{EFEFEF}\textbf{63.22}} &
		\textbf{77.47} &
		\multicolumn{1}{c|}{\cellcolor[HTML]{EFEFEF}\textbf{67.54}} &
		\textbf{80.62} &
		\multicolumn{1}{c|}{\cellcolor[HTML]{EFEFEF}{\ul 83.16}} &
		{\ul 90.8} &
		\multicolumn{1}{c|}{\cellcolor[HTML]{EFEFEF}\textbf{76.3}} &
		\textbf{86.56} &
		\multicolumn{1}{c|}{\cellcolor[HTML]{EFEFEF}\textbf{61.94}} &
		\textbf{71.97}
	\end{tabular}
\end{table*}

\begin{table*}[]
	\caption{Cross-domain RS image semantic segmentation results from  Vaihingen IRRG to Potsdam RGB.}
	\label{tab:VAI_IRRG_2_POT_RGB}
	\begin{tabular}{c|cc|cc|cc|cc|cc|cc|cc}
		\hline
		&
		\multicolumn{2}{c|}{Clutter} &
		\multicolumn{2}{c|}{Car} &
		\multicolumn{2}{c|}{Tree} &
		\multicolumn{2}{c|}{Low Vegetation} &
		\multicolumn{2}{c|}{Building} &
		\multicolumn{2}{c|}{Impervious Surface} &
		\multicolumn{2}{c}{Overall} \\ \cline{2-15} 
		\multirow{-2}{*}{Method} &
		\multicolumn{1}{c|}{IoU} &
		F1 &
		\multicolumn{1}{c|}{IoU} &
		F1 &
		\multicolumn{1}{c|}{IoU} &
		F1 &
		\multicolumn{1}{c|}{IoU} &
		F1 &
		\multicolumn{1}{c|}{IoU} &
		F1 &
		\multicolumn{1}{c|}{IoU} &
		F1 &
		\multicolumn{1}{c|}{mIoU} &
		mF1 \\ \hline
		Segformer \cite{xie2021} &
		\multicolumn{1}{c|}{2.36} &
		4.61 &
		\multicolumn{1}{c|}{\textbf{72.16}} &
		\textbf{83.83} &
		\multicolumn{1}{c|}{5.38} &
		10.21 &
		\multicolumn{1}{c|}{31.52} &
		48.65 &
		\multicolumn{1}{c|}{72.61} &
		84.13 &
		\multicolumn{1}{c|}{62.45} &
		76.89 &
		\multicolumn{1}{c|}{41.08} &
		51.39 \\
		\rowcolor[HTML]{EFEFEF} 
		AdaptSegNet \cite{tsai2018} &
		\multicolumn{1}{c|}{\cellcolor[HTML]{EFEFEF}6.11} &
		11.5 &
		\multicolumn{1}{c|}{\cellcolor[HTML]{EFEFEF}42.31} &
		55.95 &
		\multicolumn{1}{c|}{\cellcolor[HTML]{EFEFEF}30.71} &
		45.51 &
		\multicolumn{1}{c|}{\cellcolor[HTML]{EFEFEF}15.1} &
		25.81 &
		\multicolumn{1}{c|}{\cellcolor[HTML]{EFEFEF}54.25} &
		70.31 &
		\multicolumn{1}{c|}{\cellcolor[HTML]{EFEFEF}37.66} &
		59.55 &
		\multicolumn{1}{c|}{\cellcolor[HTML]{EFEFEF}31.02} &
		44.75 \\
		ProDA \cite{zhang2021}&
		\multicolumn{1}{c|}{{\ul 11.13}} &
		{\ul 20.51} &
		\multicolumn{1}{c|}{41.21} &
		59.27 &
		\multicolumn{1}{c|}{30.56} &
		46.91 &
		\multicolumn{1}{c|}{35.84} &
		52.75 &
		\multicolumn{1}{c|}{46.37} &
		63.06 &
		\multicolumn{1}{c|}{44.77} &
		62.03 &
		\multicolumn{1}{c|}{34.98} &
		50.76 \\
		\rowcolor[HTML]{EFEFEF} 
		DualGAN \cite{li2021}&
		\multicolumn{1}{c|}{\cellcolor[HTML]{EFEFEF}\textbf{13.56}} &
		\textbf{23.84} &
		\multicolumn{1}{c|}{\cellcolor[HTML]{EFEFEF}39.71} &
		56.84 &
		\multicolumn{1}{c|}{\cellcolor[HTML]{EFEFEF}25.8} &
		40.97 &
		\multicolumn{1}{c|}{\cellcolor[HTML]{EFEFEF}41.73} &
		58.87 &
		\multicolumn{1}{c|}{\cellcolor[HTML]{EFEFEF}59.01} &
		74.22 &
		\multicolumn{1}{c|}{\cellcolor[HTML]{EFEFEF}45.96} &
		62.97 &
		\multicolumn{1}{c|}{\cellcolor[HTML]{EFEFEF}37.63} &
		52.95 \\
		CIA-UDA \cite{ni2023}&
		\multicolumn{1}{c|}{9.2} &
		16.86 &
		\multicolumn{1}{c|}{63.36} &
		77.57 &
		\multicolumn{1}{c|}{44.9} &
		61.97 &
		\multicolumn{1}{c|}{43.96} &
		61.07 &
		\multicolumn{1}{c|}{70.48} &
		82.68 &
		\multicolumn{1}{c|}{53.39} &
		69.61 &
		\multicolumn{1}{c|}{47.55} &
		61.63 \\
		\rowcolor[HTML]{EFEFEF} 
		DNT \cite{chen2022a}&
		\multicolumn{1}{c|}{\cellcolor[HTML]{EFEFEF}8.43} &
		15.55 &
		\multicolumn{1}{c|}{\cellcolor[HTML]{EFEFEF}46.78} &
		63.74 &
		\multicolumn{1}{c|}{\cellcolor[HTML]{EFEFEF}36.56} &
		53.55 &
		\multicolumn{1}{c|}{\cellcolor[HTML]{EFEFEF}30.59} &
		46.85 &
		\multicolumn{1}{c|}{\cellcolor[HTML]{EFEFEF}69.95} &
		82.32 &
		\multicolumn{1}{c|}{\cellcolor[HTML]{EFEFEF}56.41} &
		72.13 &
		\multicolumn{1}{c|}{\cellcolor[HTML]{EFEFEF}41.45} &
		55.69 \\
		SiamSeg w./o. C.L. &
		\multicolumn{1}{c|}{6.99} &
		13.07 &
		\multicolumn{1}{c|}{{\ul 67.75}} &
		{\ul 80.77} &
		\multicolumn{1}{c|}{{\ul 55.82}} &
		{\ul 71.65} &
		\multicolumn{1}{c|}{{\ul 51.72}} &
		{\ul 68.17} &
		\multicolumn{1}{c|}{{\ul 79.03}} &
		{\ul 88.29} &
		\multicolumn{1}{c|}{{\ul 65.46}} &
		{\ul 79.12} &
		\multicolumn{1}{c|}{{\ul 54.46}} &
		{\ul 66.85} \\
		\rowcolor[HTML]{EFEFEF} 
		SiamSeg &
		\multicolumn{1}{c|}{\cellcolor[HTML]{EFEFEF}6.69} &
		12.55 &
		\multicolumn{1}{c|}{\cellcolor[HTML]{EFEFEF}66.76} &
		80.07 &
		\multicolumn{1}{c|}{\cellcolor[HTML]{EFEFEF}\textbf{57.96}} &
		\textbf{73.39} &
		\multicolumn{1}{c|}{\cellcolor[HTML]{EFEFEF}\textbf{53.34}} &
		\textbf{69.57} &
		\multicolumn{1}{c|}{\cellcolor[HTML]{EFEFEF}\textbf{81.52}} &
		\textbf{89.82} &
		\multicolumn{1}{c|}{\cellcolor[HTML]{EFEFEF}\textbf{68.15}} &
		\textbf{81.06} &
		\multicolumn{1}{c|}{\cellcolor[HTML]{EFEFEF}\textbf{55.74}} &
		\textbf{67.74}
	\end{tabular}
\end{table*}

\paragraph{Cross-domain RS image semantic segmentation on LoveDA Rural to Urban}

In Section \ref{dataset}, we established a cross-domain remote sensing image semantic segmentation task using the LoveDA dataset to validate the effectiveness of the proposed SimSeg method. In this section, we selected several representative comparative methods, including AdaptSegNet \cite{tsai2018}, FADA \cite{wang2020a}, CLAN \cite{luo2019}, PyCDA \cite{lian2019}, CBST \cite{zou2020}, IAST \cite{mei2020}, and DCA \cite{wu2022}.

The experimental results are presented in Table \ref{tab:Rural_2_Urban}. 
LoveDA dataset has more number of classifications and more differences between rural and urban,so most of the methods have low IoU on this dataset. Adversarial learning based methods, such as AdaptSegNet and CLAN, all perform very poorly because they cannot reduce the domain gap well. While methods using ST, such as CBST, IAST and SiamSeg w./o. C.L. have good performance, which suggests that ST is a good choice for solving cross-domain problems. However, when facing the more complex cross-domain remote sensing images, the ST method is not able to fully learn the features of the target domain due to the large domain gap, while the siamseg with the introduction of contrastive learning learns the features of the target domain better thanks to the contrastive learning, and thus enhances the performance.

\begin{table*}[]
	\center
	\caption{ross-domain RS image semantic segmentation results from LoveDA rural to urban.}
	\label{tab:Rural_2_Urban}
	\begin{tabular}{c|c|c|c|c|c|c|c|c}
		\hline
		& Agricultural & Forest & Barren & Water          & Road        & Building & Background     & Overall \\ \cline{2-9} 
		\multirow{-2}{*}{Method} & IoU          & IoU    & IoU    & IoU            & IoU         & IoU      & IoU            & mIoU    \\ \hline
		AdaptSegNet \cite{tsai2018}& 22.05        & 28.7   & 13.62  & 81.95          & 15.61       & 23.73    & 42.35          & 32.68   \\
		\rowcolor[HTML]{EFEFEF} 
		FADA \cite{wang2020a}     & 24.79        & 32.76  & 12.7   & 80.37          & 12.76       & 12.62    & 43.89          & 31.41   \\
		CLAN \cite{luo2019}       & 25.8         & 30.44  & 13.71  & 79.25          & 13.75       & 25.42    & 43.41          & 33.11   \\
		\rowcolor[HTML]{EFEFEF} 
		PyCDA  \cite{lian2019}   & 11.39        & 40.39  & 7.11   & 74.87          & 45.51       & 35.86    & 38.04          & 36.25   \\
		CBST \cite{zou2020}      & 30.05        & 29.69  & 19.18  & 80.05          & 35.79       & 46.1     & 48.37          & 41.32   \\
		\rowcolor[HTML]{EFEFEF} 
		IAST \cite{mei2020}      & 36.5         & 31.77  & 20.29  & \textbf{86.01} & 28.73       & 31.51    & \textbf{48.57} & 40.48   \\
		DCA  \cite{wu2022}       & 36.92        & {\ul42.93}  & 16.7   & {\ul 80.88}    & {\ul 51.65} & 49.6     & 45.82          & 46.36   \\ 
		\rowcolor[HTML]{EFEFEF} 
		SiamSeg w./o. C.L. & \textbf{49.23} & 42.44 & {\ul 41.73}    & 66.33 & 50.65          & {\ul 51.2}     & 36.55       & {\ul 48.3}     \\
		SiamSeg            & {\ul 49.1}     & \textbf{47.3}        & \textbf{47.44} & 71.26 & \textbf{56.67} & \textbf{52.58} & {\ul 37.65} & \textbf{51.72}
	\end{tabular}
\end{table*}

\subsubsection{Visualization Results}

In this section, we further visualize the model's predictions to validate the outstanding performance of SiamSeg. Additionally, to explore the effectiveness of contrastive learning (C.L.), we visualize and compare the predictions of SiamSeg without C.L. (SiamSeg w./o. C.L.).

\begin{figure}[!t]
	\centering
	\includegraphics[width=1.0\linewidth]{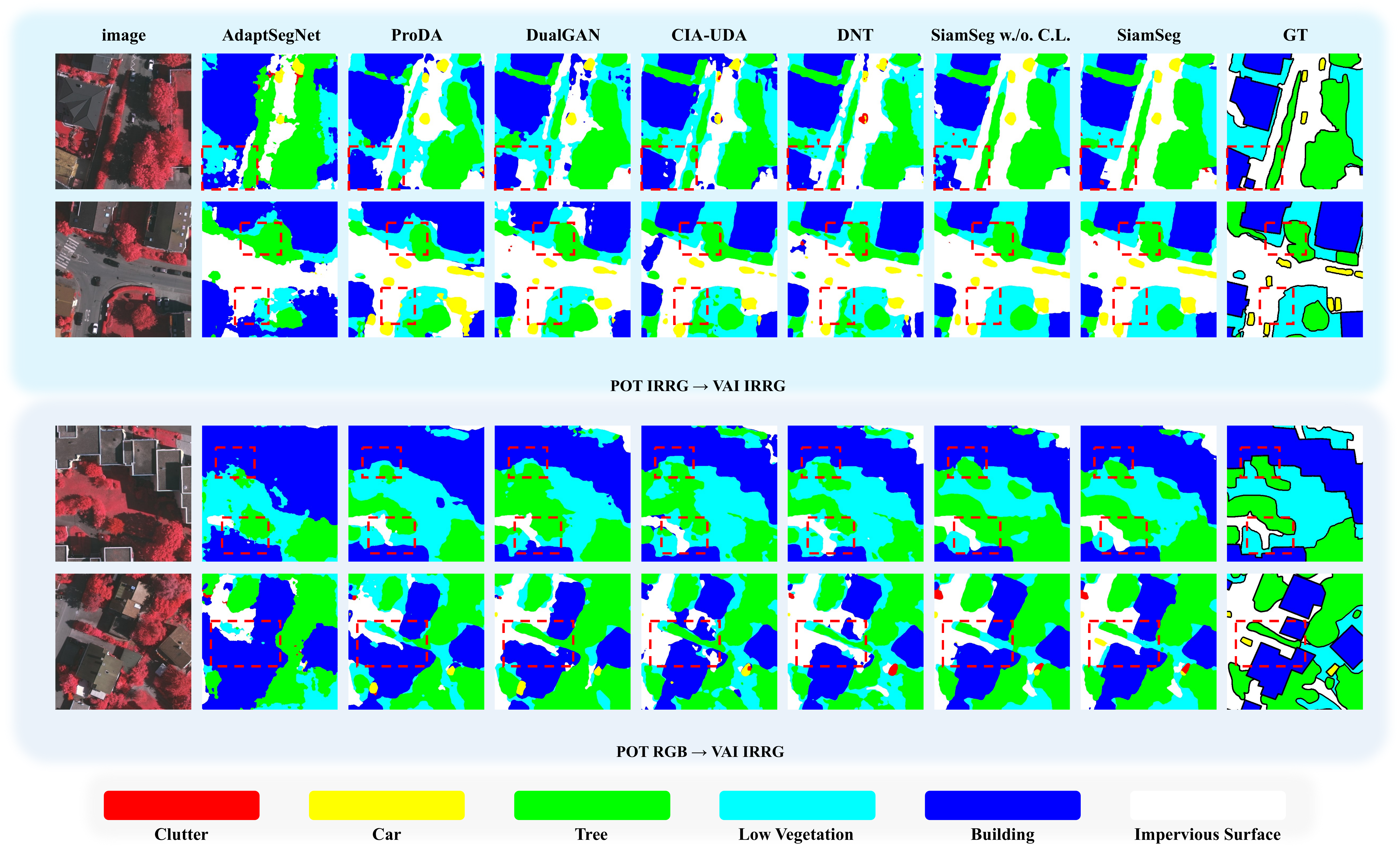}
	\caption{Visualization of results on Potsdam and Vaihingen datasets. 
	The cross-domain tasks from top to bottom are Potsdam IRRG to Vaihingen IRRG and Potsdam RGB to Vaihingen IRRG.
	The categories represented by the different colors are listed at the bottom of the picture with their names and colors.
	}
	\label{fig:POT_and_VAI_Res}
\end{figure}

The overall segmentation results, as shown in Fig. \ref{fig:POT_and_VAI_Res},  for AdaptSegNet, ProDA, and DualGAN, which are based on adversarial learning, are average due to the inability of these methods to be specially designed to account for the characteristics of a large domain gap in the RS domain. In contrast, the methods proposed by The CIA-UDA and DNT methods demonstrate excellent performance in RS image segmentation, indicating that the segmentation capabilities of a network can be significantly enhanced by addressing the challenge of inadequate learning in the target domain for RS cross-domain images. It is noteworthy that SiamSeg w./o. C.L. can achieve competitive results using only the ST method. However, due to the insufficient learning of the features in the target domain, there is considerable scope for improvement in performance. although SiamSeg w./o. C.L. achieves good segmentation results, focusing on the dotted box in the result graph, we can see that the prediction of the texture of the object and the correct classification are not as good as that of SiamSeg with the addition of contrastive learning, which shows that the addition of contrastive learning has reduced the domain gap between different domains. This also shows that the addition of contrastive learning reduces the domain gap between different domains by better learning the graph of the target domain. 

This phenomenon further corroborates the exceptional generalization capability of SiamSeg in handling cross-domain remote sensing image tasks, as well as the effectiveness of the contrastive learning method in complex scenarios.

\begin{figure}[!t]
	\centering
	\includegraphics[width=1.0\linewidth]{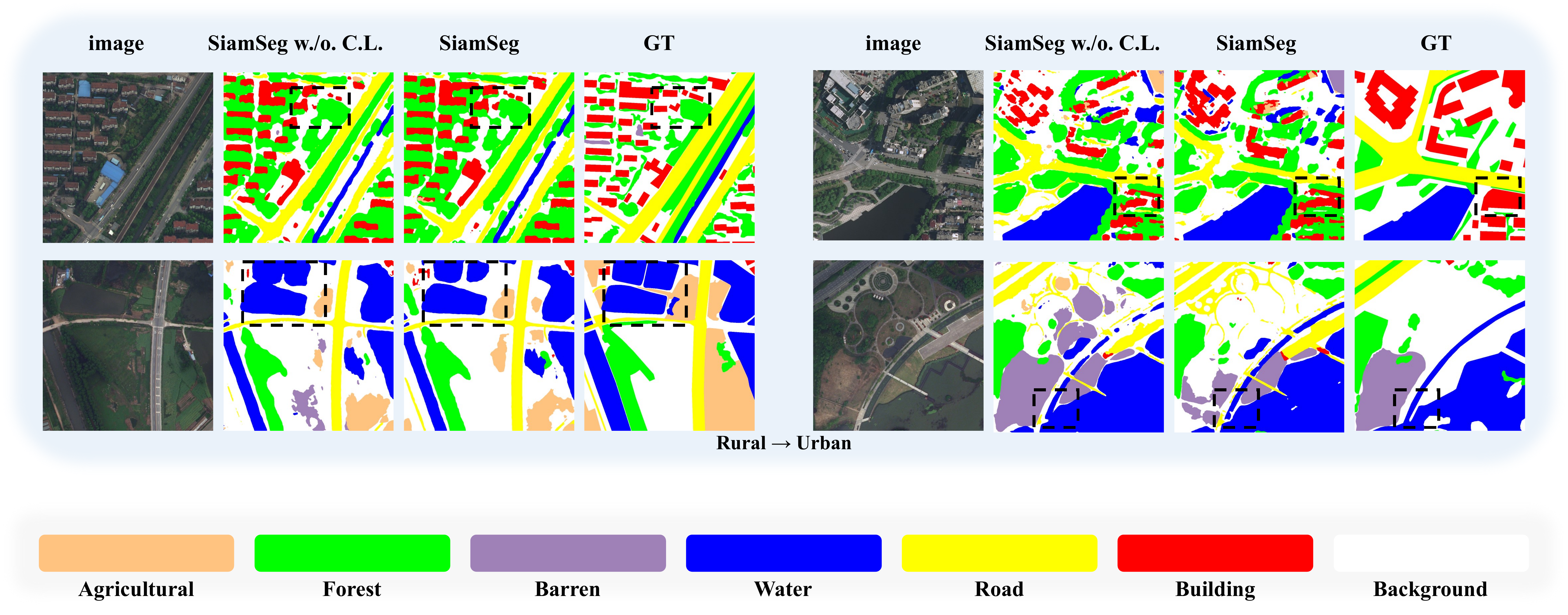}
	\caption{Visualization of results on LoveDA datasets. We conduct one task which is Rural to Urban. We provide the visualization results on LoveDA dataset. Since images in the testing dataset do not have annotations, we display the results of images in the validation dataset. 
	}
	\label{fig:LoveDA_Res}
\end{figure}

\subsubsection{Ablation Studies}
\paragraph{Effectiveness of Contrastive Learning}
Using the performance of SiamSeg and SiamSeg without Contrastive Learning (C.L.) in the task from Potsdam IRRG to Vaihingen IRRG, as shown in Table \ref{tab:Rural_2_Urban}, and visualization result in Fig. \ref{fig:LoveDA_Res}. We can clearly observe that SiamSeg outperforms SiamSeg w.o. C.L. across all categories after the introduction of C.L. This indicates that C.L. effectively enhances the model's ability, in target domain, to perceive features across different categories. The additional supervisory signals provided by C.L. enable the model to learn image features more profoundly, significantly improving performance in cross-domain semantic segmentation tasks. The introduction of C.L. enriches the feature representation of the model, thereby enhancing its generalization capabilities and category differentiation. Solved the case of large domain gap in cross-domain RS image domains.
\paragraph{Choice of Sim. Aug. Methods in Contrastive Learning}
Within the Contrastive Learning framework, we further investigated the impact of various data augmentation methods on model performance, as shown in Table \ref{tab:ablation_aug}, including Resize, Flip and Color Jitter. The experimental results indicate that Resize and Flip have a minimal effect on model performance, showing almost no significant differences. In contrast, the inclusion of Color Jitter resulted in a noticeable improvement in model performance. This may be attributed to the fact that Resize and Flip do not substantially alter the overall distribution of the images, preventing Contrastive Learning from extracting valuable representation information. In contrast, Color Jitter modifies the color features of the images, disrupting their original distribution, which allows Contrastive Learning to better learn useful representation information from similar images. Even without employing further data augmentation techniques such as Resize and Flip, Color Jitter alone significantly enhances model performance.

\begin{table}[]
	\center
	\caption{The effect of different augmentation methods. SiamSeg w. Resize denotes the version of Siamseg with the Resize augment applied in contrastive learning. C.J. denotes Collor Jitter. The task chosen in the table is POT IRRG → VAI IRRG (Potsdam IRRG to Vaihingen IRRG).}
	\label{tab:ablation_aug}
	\begin{tabular}{c|c|c|c}
		\hline
		Method            & mIoU  & mF1   & Performance \\ \hline
		SiamSeg w. Resize & 63.28 & 74.65 & ×      \\
		\rowcolor[HTML]{EFEFEF} 
		SiamSeg w. Flip   & 63.01 & 74.33 & ×      \\
		SiamSeg w. C.J.   & 65.21 & 76.16 &  \checkmark          \\
		\rowcolor[HTML]{EFEFEF} 
		SiamSeg           & 65.55 & 77.07 &  \checkmark        
	\end{tabular}
\end{table}

\section{Discussion}
\subsection{Limitations}
While the method proposed in this paper effectively addresses the performance degradation caused by unlabeled target domain data, its primary limitation lies in its reliance on joint training with labeled source domain data and unlabeled target domain data. This necessitates the separate training of a model for each cross-domain task, increasing training costs. Furthermore, due to the dependency on source domain data, training the domain adaptation model can be time-consuming, especially when computational resources are limited. When the volume of source domain data is substantial, retraining a model for each new task becomes impractical. Nevertheless, it is important to emphasize that the proposed method has a minimal dependence on target domain labels, providing a significant advantage in unsupervised scenarios. Compared to traditional methods that rely on target domain labels, our approach demonstrates greater adaptability in unlabeled environments.
\subsection{Future Works}
Future research will further explore Source-Free Domain Adaptation (SFDA) methods, particularly in scenarios where source domain data is unavailable, to achieve more effective migration without source domain data. SFDA methods eliminate the need to reuse source domain data for training in each task, aligning more closely with practical application requirements, especially in cases where source domain data is difficult to obtain or usage is restricted. Additionally, we will focus on enhancing the model's generalization ability, enabling it to adapt efficiently across different cross-domain tasks, thereby further reducing reliance on source domain data, lowering training costs, and increasing application flexibility.
\section{Conclusion}
This paper presents the SiamSeg method, which effectively addresses domain migration issues in remote sensing image cross-domain semantic segmentation tasks by integrating contrastive learning. This approach not only enhances the model's perceptual capability for target domain features through unsupervised learning but also significantly improves the model's cross-domain generalization ability, particularly excelling in recognizing complex categories such as buildings and roads. Experimental results demonstrate that SiamSeg achieves higher mean Intersection over Union (mIoU) and accuracy compared to existing methods, while maintaining a low computational complexity. Future work will continue to focus on reducing dependence on source domain data and exploring more efficient domain adaptation techniques to further enhance the practical application value of the model.

\bibliographystyle{IEEEtran}
\bibliography{sample}
\vspace{-15 mm} 
\begin{IEEEbiography}[{\includegraphics[width=1in,height=1.25in,clip,keepaspectratio]{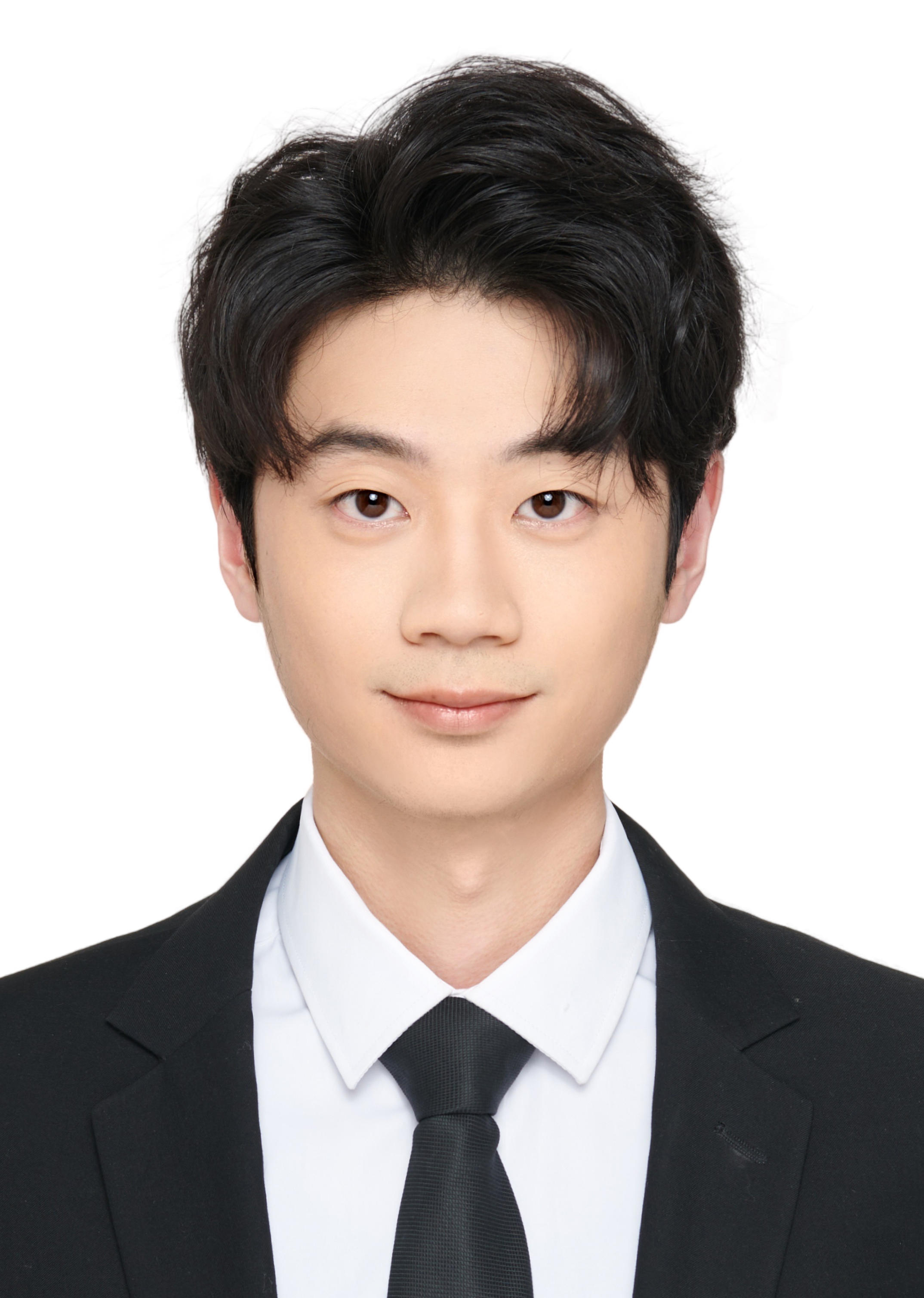}}]{Bin Wang}
	received the B.Sc. degree from the Chengdu University of Technology (CDUT), College of Computer Science and Cyber Security, Chengdu, China, in 2022, where he is pursuing the M.Sc. degree in computer science and technology. His research interests include intelligent geophysical data processing, computer vision and applications of deep learning.
\end{IEEEbiography}
\vspace{-15 mm} 
\begin{IEEEbiography}[{\includegraphics[width=1in,height=1.25in,clip,keepaspectratio]{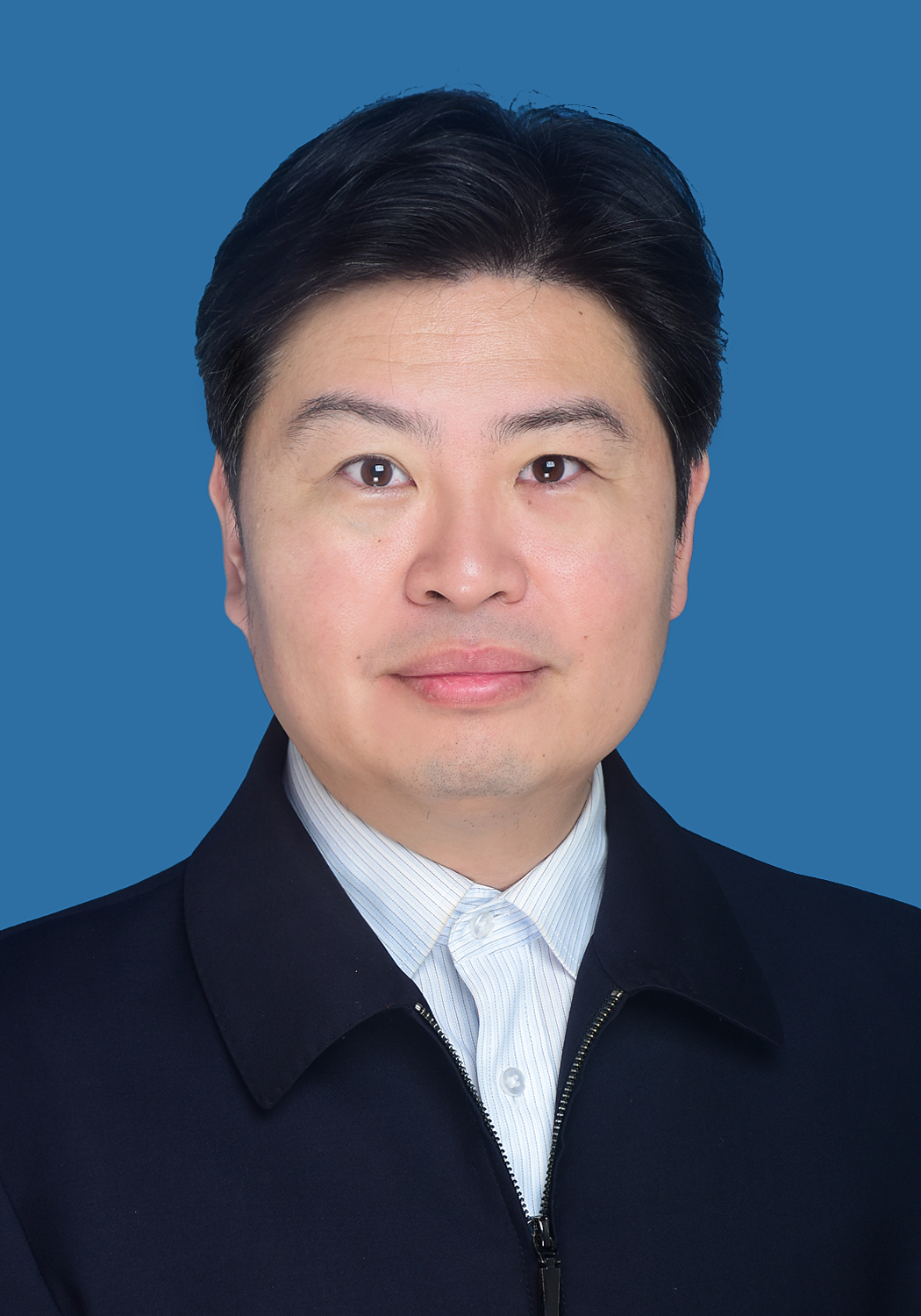}}]{Fei Deng}
	received the Ph.D. degree in Earth exploration and information technology from the College of Information Engineering, Chengdu University of Technology, Chengdu, China, in 2007.
	Since 2004, he has been with the College of Computer and Network Security, Chengdu University of Technology, where he is currently a Professor. His research interests include artificial intelligence, deep learning, and computer graphics.
\end{IEEEbiography}
\vspace{-15 mm} 
\begin{IEEEbiography}[{\includegraphics[width=1in,height=1.25in,clip,keepaspectratio]{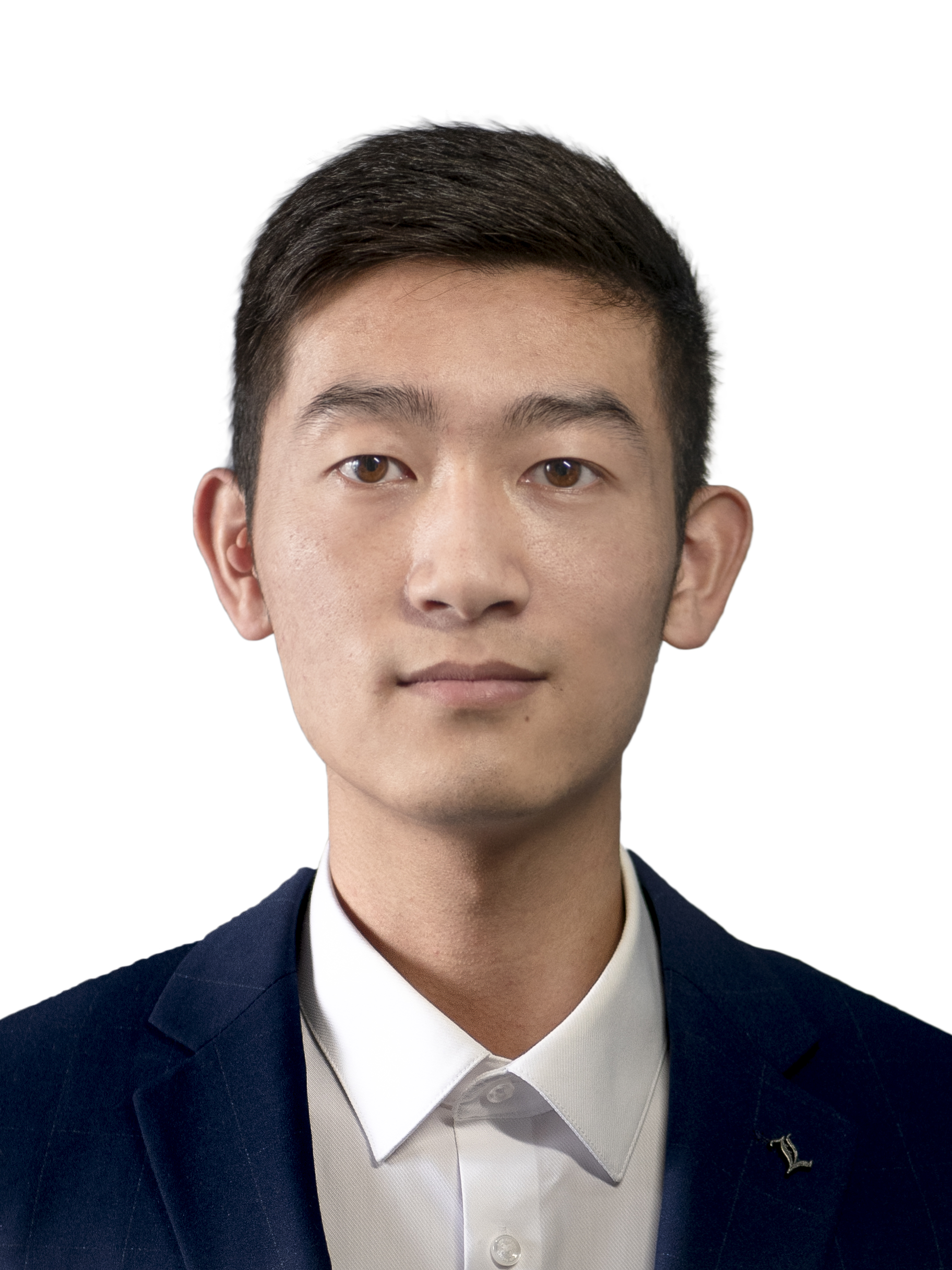}}]{Shuang Wang}
	received the B.S. degree in software engineering from Chengdu University of Technology, Chengdu, China, in 2022, where he is currently pursuing the Ph.D. degree in Earth exploration and information technology. His research interests include applications of deep learning, computer vision, complex signal processing, and intelligent geophysical data processing and
	modeling.
\end{IEEEbiography}
\vspace{-15 mm} 
\begin{IEEEbiography}[{\includegraphics[width=1in,height=1.25in,clip,keepaspectratio]{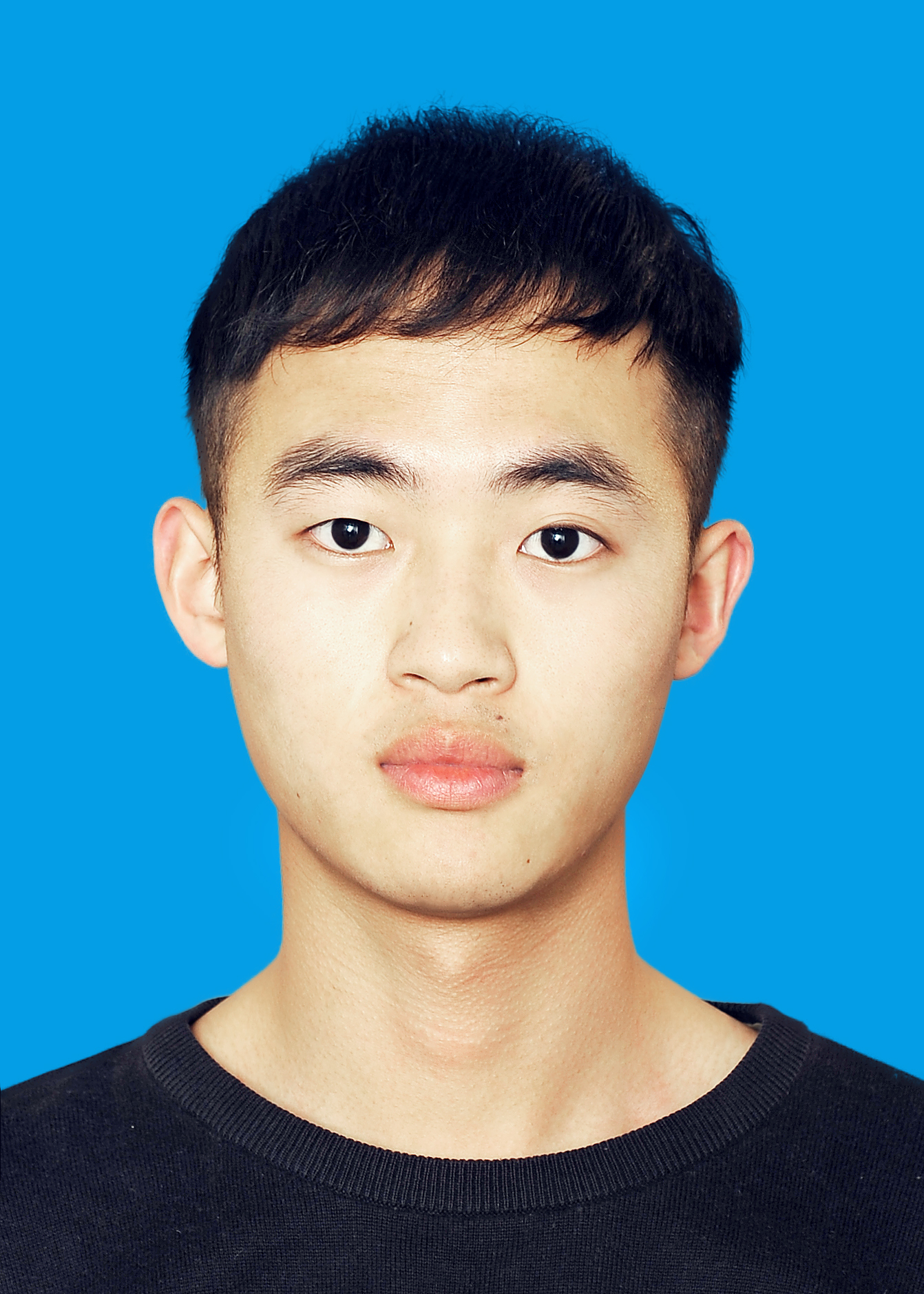}}]{Wen Luo, Member, IEEE}
	received the Master of Electronic Information degree in computer technology from Chengdu University of Technology, Chengdu, China, in 2024, where he is currently pursuing the Ph.D. degree in geological resources and geological engineering.
	His research interests include artificial intelligence, computer vision, and remote sensing image recognition.
\end{IEEEbiography}
\vspace{-15 mm} 
\begin{IEEEbiography}[{\includegraphics[width=1in,height=1.25in,clip,keepaspectratio]{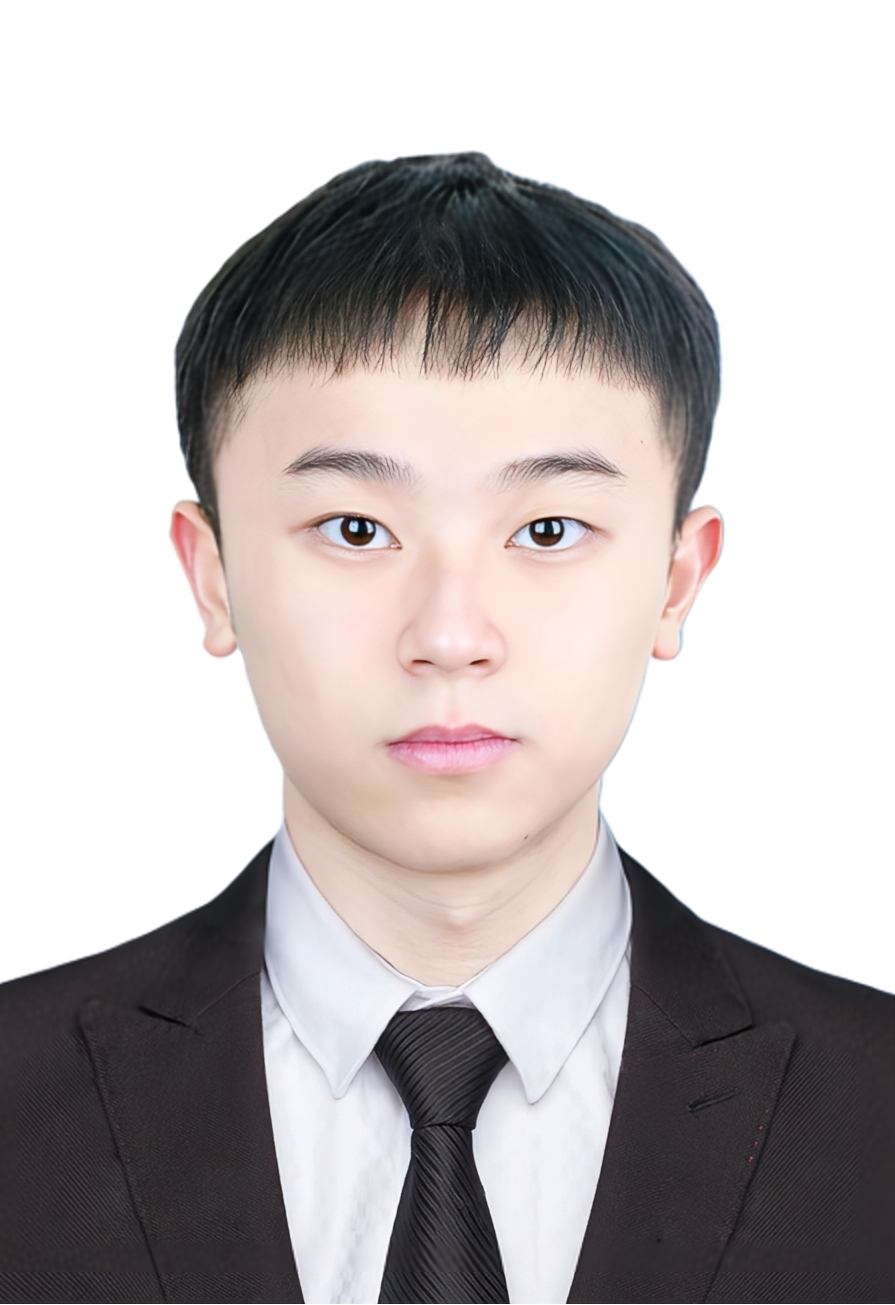}}]{Zhixuan Zhang}
	He is currently pursuing the B.E. degree in Intelligent Manufacturing Engineering from College of Mechanical and Vehicle Engineering, Changchun University, Changchun, China. His research interests include artificial intelligence, computer vision, medical image processing.
\end{IEEEbiography}
\vspace{-15 mm} 
\vspace{-15 mm} 
\begin{IEEEbiography}[{\includegraphics[width=1in,height=1.25in,clip,keepaspectratio]{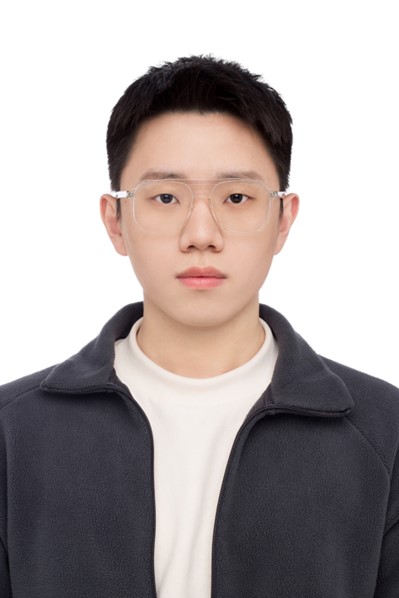}}]{Peifan Jiang, Graduate Student Member, IEEE}
	received the M.S. degree in computer technology from Chengdu University of Technology, Chengdu, China, in 2023, where he is currently pursuing the Ph.D. degree in Earth exploration and information technology. His research interests include applications of deep learning and intelligent geophysical data processing and modeling.
\end{IEEEbiography}
\end{document}